\documentclass[sigconf, nonacm]{acmart}
\setcopyright{none}
\graphicspath{{fig/}{./}}

\usepackage{bbold}
\usepackage{enumitem}
\usepackage{multirow}
\usepackage{subcaption}
\usepackage{xcolor}
\usepackage{xspace}
\usepackage{adjustbox} 
\usepackage{booktabs}
\usepackage{tcolorbox}
\tcbuselibrary{breakable}
\tcbset{breakable}
\usepackage{geometry}
\usepackage{listings}
\usepackage{xcolor}
\usepackage{makecell}

\newtcolorbox{promptbox}[2][]{width=\linewidth,
boxsep=2pt,left=8pt,right=7pt,top=5pt,bottom=5pt,
fontupper=\ttfamily,fontlower=\ttfamily,
fonttitle=\hypersetup{linkcolor=white,urlcolor=white},
title={#2},
label={#1}
}
\newtcolorbox{exbox}[2][]{width=\linewidth,
boxsep=1pt,left=2pt,right=2pt,top=0pt,bottom=0pt,
fontupper=\ttfamily,fontlower=\ttfamily,
fonttitle=\hypersetup{linkcolor=white,urlcolor=white},
title={#2},
label={#1}
}

\newcommand{\todo}[1]{\textcolor{red}{TODO: #1}}

\newcommand{\name}{VeriLA}
\begin{document}

\title{\name{}: A Human-Centered Evaluation Framework for \\Interpretable Verification of LLM Agent Failures}

\author{Yoo Yeon Sung\textsuperscript{\textdagger}}
\affiliation{%
  \institution{University of Maryland}
  \country{USA}
}
\email{yysung53@umd.edu}

\author{Hannah Kim}
\affiliation{%
  \institution{Megagon Labs}
  \country{USA}
}
\email{hannah@megagon.ai}

\author{Dan Zhang}
\affiliation{%
  \institution{Megagon Labs}
  \country{USA}
}
\email{dan_z@megagon.ai}

\thanks{\textdagger~Work done during an internship at Megagon Labs.}

\renewcommand{\shortauthors}{Sung et al.}

\begin{abstract}
AI practitioners increasingly use large language model (LLM) agents in compound AI systems to solve complex reasoning tasks, these agent executions often fail to meet human standards, leading to errors that compromise the system's overall performance. 
Addressing these failures through human intervention is challenging due to the agents' opaque reasoning processes, misalignment with human expectations, the complexity of agent dependencies, and the high cost of manual inspection.
This paper thus introduces a human-centered evaluation framework for \textbf{V}erifying \textbf{L}LM \textbf{A}gent failures (\name{}), which 
systematically assesses agent failures to reduce human effort and make these agent failures interpretable to humans.
The framework first defines clear expectations of each agent by curating human-designed agent criteria. 
Then, it develops a human-aligned agent verifier module, trained with human gold standards, to assess each agent's execution output.
This approach enables granular evaluation of each agent's performance by revealing failures from a human standard, offering clear guidelines for revision, and reducing human cognitive load. Our case study results show that \name{} is both interpretable and efficient in helping practitioners interact more effectively with the system. 
By upholding accountability in human-agent collaboration, \name{} paves the way for more trustworthy and human-aligned compound AI systems.
\end{abstract}



%   \includegraphics[width=\linewidth]{fig/abstract_re.pdf}
%   \caption{Overview of \name{}. Our pipeline consists of 1) task decomposition by an LLM planner, 2) execution of subtasks by LLM executors, and 3) verification of subtask outputs by a trained verifier.}
%   \Description{\todo{add}}
%   \label{fig:diagram}


\maketitle
\section{Introduction}
\label{sec:intro}
\begin{figure}
  \includegraphics[width=\linewidth]{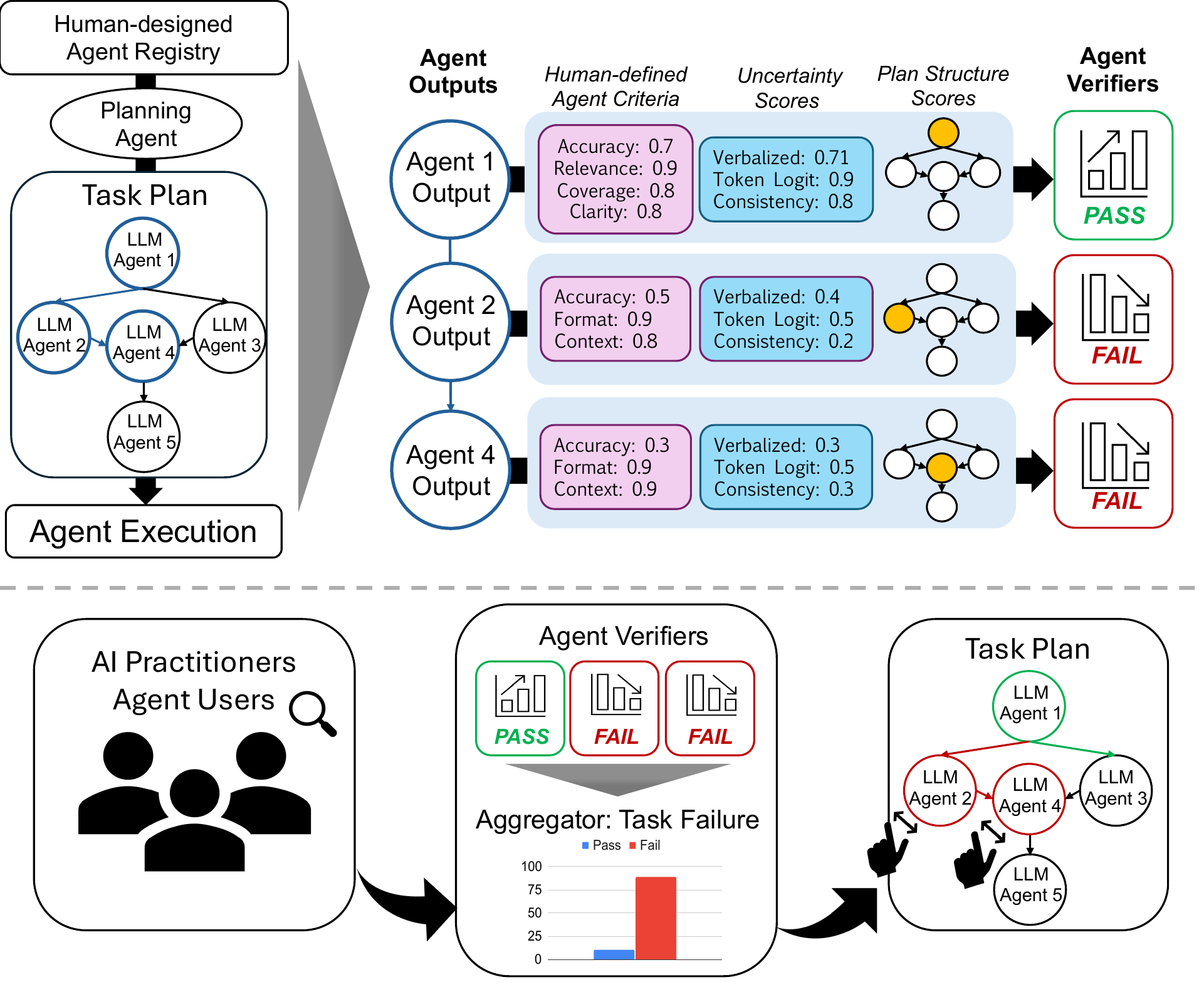}
    \caption{Overview of \name{}. Our framework operates in three main stages (1) planning where a planning agent decomposes a task into subtasks using a human-designed agent registry and generates a plan graph; (2) agent execution where specialized LLM agents perform the subtasks; and (3) execution verification, which verifies each LLM agent's outputs based on human-defined agent criteria, agent uncertainty, and dependency information from the plan structure. We then assess task failure with aggregation metrics that combine verifier scores. Our framework guides users to detect task failures efficiently, identify faulty agents, and analyze the root causes of their failure. }
  \label{fig:diagram}
\end{figure}

%As large language models (LLMs) continue to excel across various fields, they are increasingly utilized to address real-world complex tasks through LLM-based systems~\cite{suzgun2022challenging}. In particular, the concept of LLM-as-agent systems has emerged, which are used in approaches like task decomposition or chaining to tackle these challenges just like humans~\cite{srivastava2023beyond, wu2022aichains}. 

% However, these agent-based systems face significant limitations in handling more complex reasoning tasks and often contradict human cognition, often requiring human intervention for feedback and self-revision~\citep{langchain, arawjo2024chainforge}. 
%This places a significant burden on humans, who must review each agent's execution inputs and outputs while grappling with the ambiguity of structurally assessing whether the execution is correct. 
As large language models (LLMs) continue to excel across various fields, they are increasingly used to address complex reasoning tasks through LLM-as-agent systems~\citep{xi2025rise, wang2024survey}. A key application is a LLM-based compound AI system, where a planning agent breaks a complex task into simpler subtasks, and delegates these subtasks to multiple specialized LLM agents~\cite{wu2022aichains, zhang2023building, compound-ai-blog}. Each assigned agent must accurately execute its subtask, as the final agent's output heavily relies on the previous agents' outputs and is considered as the final solution of the overall task~\citep{cheng2024exploring}. Failures in any agent can propagate and cause the overall task
 failure~\citep{sumers2023cognitive, jaeger2013uncertainty}.
% bigbench citation: srivastava2023beyond
%
While these compound systems demonstrate strong problem-solving capabilities, they face significant limitations in that they may produce outputs that contradict human expectations, often requiring human intervention for failure feedback and revision~\citep{langchain, arawjo2024chainforge}.
However, providing feedback is challenging because agent outputs come from reasoning that deviates from humans or lack clarity on the cause of failure, hindering users from providing guidance on remedying execution failures.
%this poor interpretability hinders users from providing effective guidance for revision, especially when pinpointing failures within individual agent is crucial, as such errors can propagate and result in overall task failure. 
%
Moreover, manually reviewing each step is labor intensive and not scalable. It forces users to audit each agent’s execution outputs, increasing the risk of errors.
% This process increases the risk of errors and subjective judgments, resulting in less-structured decisions due to ungrounded reasoning. 
%struggling with the ambiguity of structurally determining correctness.
This underscores the need for more systematic and efficient methods for auditing and supervising compound AI systems with LLM agents.

%To address this, we propose a framework that evaluates the correctness of each agent's execution autonomously, while ensuring it aligns with human-grounded values. We take a human-aligned approach by building agent-specific evaluation modules that integrate human's standard when evaluating whether the each agent's output fulfills its assigned task. This process emulates the human metacognitive process of self-monitoring~\citep{Winne_Azevedo_2022} to validate agent's execution result~\citep{liao2023rethinking}. Moreover, this module enables the agent's self-evaluation scored based on its certainty levels under the assumption that lower certainty often leads to deviations from the set criteria. This design ensures the interpretability of the systems on which each agent is based, allowing identification of which agent has failed and why.
To address this challenge, 
our framework consists of three interconnected components: (1) \textbf{planning}, where a planning agent decomposes the target task into simpler subtasks based on a predefined agent registry, generating a graph-based plan. The agent registry provides each agent with clear role assignments and execution guidelines that adhere to human expectations; (2) \textbf{agent execution}, in which LLM agents sequentially perform their designated subtasks, with success determined by their adherence to the predefined roles; (3) \textbf{execution verification}, where a verifier module automatically assesses each agent output to ensure its success in
fulfilling its assigned subtask. Our verifier incorporates human-centered judgments on agent outputs along with its relationship with other agents: an agent's subtask type, scores based on human-defined agent criteria, agent uncertainty, and agent's dependency within the plan structure.
%By incorporating human judgment criteria, our verifier emulates the human metacognitive process of self-monitoring~\citep{Winne_Azevedo_2022} to validate execution results~\citep{liao2023rethinking}. 
% Additionally, we integrate two LLM uncertainty estimation techniques---logit-based and self-consistency-based---under the assumption that lower certainty often correlates with a higher likelihood of errors or deviations. 
%\todo{shorten below}
%We present a metric that aggregates verifier results across agents. It identifies failed tasks with agent failures, allowing users to efficiently pinpoint and prioritize problematic tasks without reviewing all failures. By incorporating the plan's graph structure, the metric enables focused analysis of specific agent verifiers and failure criteria, proving highly effective in detecting task failures.
In addition to verifying the execution of individual agents, \name{} provides additional verification of overall task success. We introduce a metric that aggregates verifier results across agents to identify failed tasks due to agent failures. By leveraging the plan's graph structure, it enables targeted analysis of verifiers and failure criteria, streamlining failure detection. To evaluate our framework, we conducted a case study on a complex reasoning task---solving mathematical reasoning problems---demonstrating its effectiveness and practical applicability to AI practitioners.

In summary, \name{} enables detailed auditing of agentic systems to ensure transparency in agent failures, reducing manual review efforts, and strengthening trust between users and compound AI systems.
Beyond merely automating the validation of each agent's success, we hope for a collaborative problem-solving between human and LLM agents by tailoring agent workflow to human needs.

\section{Related Work}
\subsection{Need for Human Intervention in Compound AI Systems}
LLMs are capable of reasoning, tool usage, planning, and following instructions~\cite{lou2024largelanguagemodelinstruction}.
For tasks that are too complex for n-shot CoT prompting, LLMs can decompose them into subtasks and delegate each subtask to several agents~\cite{huang2024understanding}. 
This decomposition externalizes LLMs' inner reasoning, allowing users to intervene at intermediate steps---such as suggesting alternatives or correcting errors---to ensure alignment with human expertise and domain-specific requirements~\citep{wu2022aichains, langchain, arawjo2024chainforge, jigsaw}.
While recent LLMs can plan and self-refine tasks~\cite{huang2024understanding}, issues such as hallucinations and limited feedback scopes remain~\cite{sun2023adaplanner}, making human intervention crucial to detect and correct errors before they propagate~\citep{grundemclaughlin2024designingllmchainsadapting}.
For example, errors such as misinterpreting a calculation or producing an inaccurate response may go unnoticed by LLM-only evaluation but can be handled by humans.
Thus, while LLMs can help with task decomposition and execution~\cite{lou2024largelanguagemodelinstruction}, the integration of human intervention remains essential for reliability and accuracy, especially in error-sensitive domains such as legal or medical applications.

\subsection{Human-Centered Evaluation of Language Models}
% Narrowing the socio-technical gap has emerged as a significant challenge within the HCI community, focusing on understanding how evaluation results should be utilized and by whom. For instance, \citet{liao2023rethinking} argues that evaluation modules must provide valid assessments of whether and to what extent human needs are met in downstream use cases, ensuring human requirement realism. Similarly, \citet{arabzadeh2024towards} emphasizes that identifying and quantifying the criteria for LLM-powered applications is essential to verify whether they satisfy user requirements and ultimately bring utility to end-users. In math reasoning tasks, the agent’s success lies not only in arriving at the correct solution but also in presenting it in a manner that meets critical criteria such as completeness, conciseness, and clarity. These factors are crucial for verifying whether the application aligns with user expectations and delivers value in real-world contexts~\citep{ibrahim2024beyond}. Kim et al. (2024) demonstrate that incorporating user-defined criteria enhances the alignment between LLM evaluator outputs and human-annotated outputs. Similarly, Shankar et al. (2024) propose an approach to align LLM evaluators with human-defined criteria, highlighting the importance of tailoring evaluations to downstream use cases and their specific requirements. Building on this line of work, we adopt human-designed criteria to evalauate execution success, which serve as key features for agent evaluation module development within our framework.
Narrowing the socio-technical gap is a key challenge in HCI, focusing on how evaluation results should be utilized and by whom. 
\citet{liao2023rethinking} argue that evaluation modules must assess how well human needs are met in downstream use cases. 
Similarly, \citet{arabzadeh2024towards} emphasize the importance of identifying criteria for LLMs. 
For example, in math reasoning tasks, an agent's success depends not only on producing correct solutions but on presenting them with completeness, conciseness, and clarity---critical criteria for such task~\citep{ibrahim2024beyond}. 
Also, user-defined criteria improve alignment between LLM evaluator outputs and human judgements~\cite{kim2024evallm}.
\citet{shankar2024validates} propose an approach to align LLM evaluators with human-defined criteria, tailoring evaluations to specific use cases. 
Building on this line of work, we adopt human-defined agent criteria to evaluate agent's execution success in our framework.

\subsection{Assessing LLM Agent's Uncertainty}
Several techniques have been proposed to assess the uncertainty level of LLMs, including methods based on token likelihoods, consistency across multiple prompts, and self reflection~\citep{chen2024quantifying}. 
A common approach leverages token likelihood, such as logit-based confidence and entropy-based confidence, which rely on the LLM's internal probability distribution over potential tokens~\citep{huang2023lookleapexploratorystudy} (all uncertainty methods detailed in Appendix~\ref{app:verifier-features-uncertainty}). 
Another approach, verbalized confidence~\citep{xiong2023can}, asks the LLM to articulate its confidence, either as qualitative indicators or explicit confidence scores. 
Uncertainty can also be assessed using external LLM evaluators, utilizing verbalized confidence from their response or logit-based confidence from its verification assessment (e.g., logits for the prediction of a "TRUE" token)~\cite{geng-etal-2024-survey}.
Additionally, self-consistency methods~\citep{wang2022self} compare outputs from multiple runs: frequency-based metrics~\citep{yona2024largelanguagemodelsfaithfully} assess agreement across outputs, while others use weighted aggregation with logit or verbalized confidence~\cite{xiong2023can}.

\begin{figure*}
  \includegraphics[width=\linewidth]{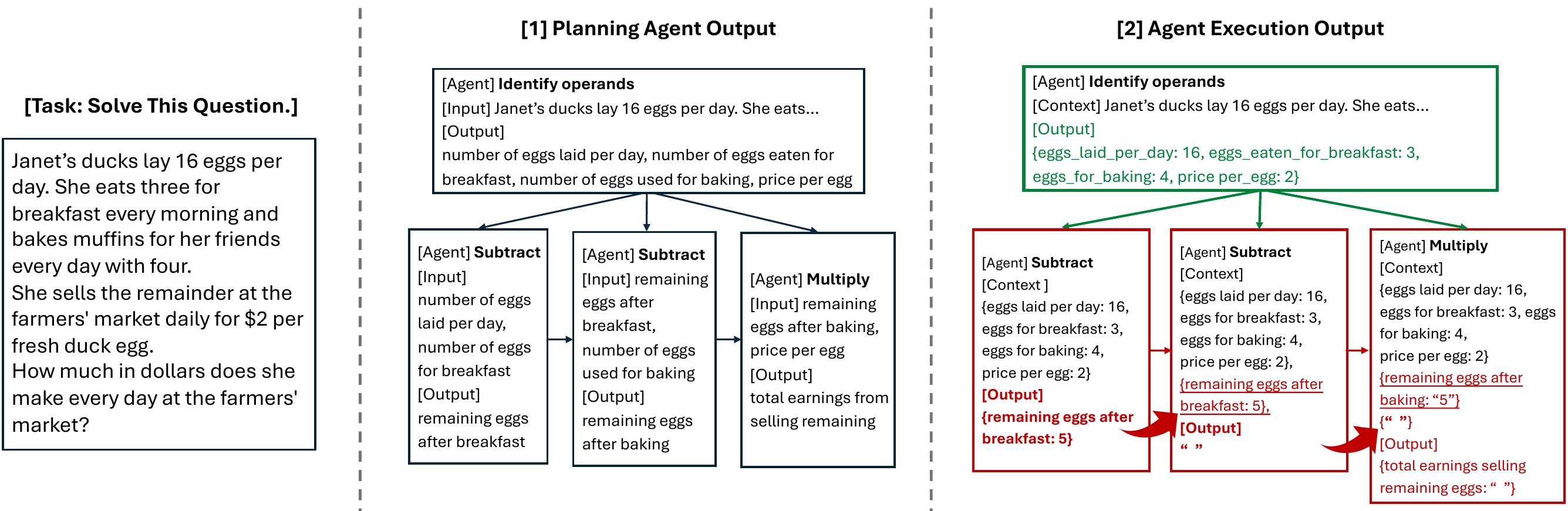}
  \caption{Example of agent's failure propagating to overall task failure. For example, based on the generated plan from the planning agent, each agent should accurately execute their subtasks. The first ``subtract'' agent failed to calculate the remaining eggs, causing subsequent ``subtract'' and ``multiply'' agents to lack the necessary context for a successful execution (three red boxes). An agent-specific verifier can help users trace the error propagation, identify the root cause of the error, and understand how it led to the task failure.}
  %\Description{\todo{add}}
  \label{fig:agent_failure_example}
\end{figure*}
\section{\name{}: Framework for Verifying LLM Agents in Compound AI Systems}
We propose an evaluation framework that aims to assist users in auditing and interacting with compound AI systems. During users' inspection of overall task failures, they can detect each agent's execution failures, and quickly and clearly understand the reasons behind the failure. This enables them to provide actionable suggestions for planning or execution revisions. 

%\subsection{Planner as task decomposer}
\subsection{Planning}
\label{sec:planning}
% To align agent outputs with human needs,
% we first curate an agent taxonomy and criteria where three AI practitioners assign each agent with specific roles and required input and output information Then, an LLM task planner, an agent that decomposes the original task into subtasks, is instructed to generate a plan in a graph structure. 
In a compound AI system, a planning agent's role is to decompose a task into a sequence of subtasks, assign them to specialized agents---which are typically predefined in the system's agent registry---and generate a plan to solve the task.\footnote{The planning agent is not registered in agent registry and thus does not participate in the plan.} 

To curate an agent registry that aligns with the human reasoning process, we first ask AI practitioners to curate a system's agent registry tailored to the target application. They register each agent with a specific role, along with its expected input and output. For example, in a math reasoning task, an ``Add'' agent is responsible for summing given operands, where the input is a list of numbers, and the output is a single sum (Table~\ref{tab:agent_registry}). \footnote{To guide practitioners on the necessary agents and their required functionalities, we use Chain-of-Thought (CoT) prompting to generate a pool of agent candidates. Then, they identify the most common agents and craft their roles, inputs, and outputs (Appendix~\ref{app:cot}).} Using this agent registry, the planning agent decomposes a given task into subtasks and delegates them to appropriate agents (Upper left in Figure~\ref{fig:diagram}). This later helps users in diagnosing each agent's failure; it informs them whether the agent was executed appropriately based on its role and has a proper output format.

Then, our planning agent generates a plan with a directed acyclic graph (DAG) format, where each node represents an agent, and directed edges show input-output dependencies between the nodes (example in Appendix~\ref{app:dag-plan}). This graph-based planning ensures that the task complexity is decomposed into an interrelated sequence of simplified subtasks. Then, each subtask is assigned to an appropriate agent with specific inputs and outputs, mirroring how humans often approach complex tasks~\citep{ghallab2004automated}.

% removing this since criteria curation is moved to 3.3
 % Next, practitioners establish carefully designed criteria for each agent to ensure that expectations align with human standards (example in Appendix~\citep{tab:app:human-criteria}). Grounding agents' evaluation in predefined criteria allows for objective judgment and a clear understanding of why an agent may have failed. (e.g., [2] EXECUTOR: \textit{subtract} agent was not accurate in calculating \textit{\small{the remaining eggs after breakfast}}). This process enables the planner to precisely assign responsibilities to each executor along with their expected outcomes so that the planner can systematically break down the task (e.g., [1] PLANNER in Figure~\ref{fig:task_decomp_pipeline}).

Planning agents also generate instruction prompts for each agent, reflecting its role and input--output format. These prompts should be closely tied to the original task, reducing the risk of agents hallucinating or forgetting the trajectory needed to solve the task.
 %
% As a result, the generated plan is represented with directed acyclic graphs (DAGs), where each node represents an agent, and directed edges show dependencies between the nodes. This graph-based decomposition ensures that each subtask receives only the necessary inputs, overcoming the limitations of linear decomposition methods used in many existing LLM-based approaches.
%
%curate a list of agents tailored to the target application, establishing a predefined agent taxonomy. Each agent is defined with a specific role, along with its expected input and output. For example, an ``Add'' agent is responsible for summing given operands, where the input is a list of numbers, and the output is a single sum.
% \hannah{We also generate subtask instruction prompts alongside a plan that reflect the expected inputs and outputs for each node, following the predefined taxonomy. These prompts should be closely tied to the original task, reducing the risk of the LLM executor hallucinating or forgetting the trajectory that should be undergone to yield the correct final output.}

%\subsection{Executor as LLM agent following assigned roles}
\subsection{Agent Execution}
% Executors, individual agents, \dz{executors of agents or executors which are agents} 
Agents execute assigned specific subtasks and instruction prompts from the generated plan. Additionally, they receive relevant context information as input which includes relevant outputs produced from the preceding agents. Because agents can fail during execution, users must intervene to correct errors and prevent an agent's failure from propagating to overall task failure. However, manually auditing whether an agent has fulfilled its subtask is both cumbersome and cognitively demanding for users. Thus, we introduce a agent-specific verifier in the next section.

%\subsection{Evaluator as assessing agents with human grounding}
\subsection{Execution Verification by Human-aligned Agent Verifier}\label{sec:verifier}
% To alleviate this burden, our evaluation results provide a filter, \dz{a bit confusing how "results provide a filter". along with global assertion, return a mask over nodes which masked out the correct subtasks?} allowing the verifier to assess subtask executions, identify potentially incorrect results, and direct user attention only to problematic intermediate outputs. 
Our agent verifier autonomously evaluates agent executions and flags potentially incorrect results, enabling users to focus only on the problematic agents within the plan.
Although some self-verifying agents exist~\citep{sun2023adaplanner, madaan2024self}, we question their reliability~\citep{stechly2024self}, lack of contextual understanding~\citep{prasad2023adapt}, and insufficient human alignment~\citep{goyal2024designing}.

To address these issues, we build a separate evaluation module for each agent to detect execution failures. This module functions as a binary classifier, trained on multiple features such as human-defined agent criteria, agent uncertainty, and plan structures with subtask types. To balance verifier's autonomy with human-designed criteria, we integrate external LLM judge scores that are assessed with human-defined criteria (Appendix~\ref{app:human-criteria}).
These criteria are carefully tailored for each agent to ensure that expectations align with human standards. This way, our verifier ensures that each agent aligns with human expectations and is evaluated according to users' specific needs~\citep{liao2023rethinking}.
For instance, the ``Subtract'' agent's execution is evaluated not only on the accuracy of summation but also on its adherence to the expected format (e.g., number) and sufficiency of context information for the subtask (Figure~\ref{fig:agent_failure_example}).
Grounding agents' evaluation in predefined criteria permits objective judgment and clarifies failure reasons.
%(e.g., [2] EXECUTOR: \textit{subtract} agent was not accurate in calculating \textit{\small{the remaining eggs after breakfast}}).
%seems to be repeating sentence from 2nd para
% To align the verifier's prediction with humans and improve its accuracy, we design our feature set to include LLM evaluation based on human-defined criteria, agent's uncertainty, and features that reflect the subtask characteristics within the plan. 
% We include external LLM evaluator scores based on human-judgmed criteria. By incorporating these criteria, our verifier emulates the human metacognitive process of self-monitoring~\citep{Winne_Azevedo_2022} to validate execution results~\citep{liao2023rethinking}.

% To estimate uncertainty, we integrate two LLM uncertainty estimation techniques—logit-based and self-consistency-based—under the assumption that lower certainty often correlates with a higher likelihood of errors or deviations (see Appendix~\ref{app:verifier-features-uncertainty} for details).

Additionally, we integrate three LLM uncertainty estimation techniques: verbalized confidence, logit-based confidence, and confidence based on self-consistency. This is under the assumption that lower certainty often correlates with a higher likelihood of errors or deviations (details on uncertainty features in Appendix~\ref{app:verifier-features-uncertainty}).
Finally, we include each agent’s subtask type and its position within the plan’s DAG structure, by using subtask categorical encoding and features like the number of preceding nodes to capture dependency relationships between agents (Appendix~\ref{sec:plan_features}). These structural features enhance our verifier’s ability to assess execution reliability within the overall task.

% included in sec 4
% We use GPT-4o for all experiments, employing a temperature setting of 0.7 for consistency-related metrics to obtain as diverse answers as possible, and a temperature of 0.1 for the remaining experiments ~\citep{xiong2023can}.

\paragraph{\textbf{Ground Truth to Train Agent Verifier}}\label{sec:verifier_label}
We use human annotated labels as ground truth to train our agent verifier. By learning execution feature patterns from these labels, the model predicts whether an agent is likely to fail during execution based on human expectations.
To ensure accurate and fair labeling, human annotators assess if the agent correctly performed its task, following the agent registry roles and human-defined criteria. Annotators are given context information, clear role definitions, expected output format, and the same human-defined criteria used by LLM judges. 

\subsection{\textbf{Verifier-Driven Aggregation Metrics for Overall Task Failure Prediction}}
% \begin{figure}
%   \includegraphics[width=\linewidth]{fig/aggregator_re.pdf}
%   \caption{}
%   \Description{\todo{add}}
%   \label{fig:diagram}
% \end{figure}
To support human-agent interaction, we guide users in identifying task failures when decomposed and executed by agents. Our verifier predicts potential failures and provides confidence scores for each agent's execution, which we use to assess overall task success.
We propose several aggregation metrics to aggregate individual verifier scores. The metric scores represent the likelihood of overall task success, with higher scores indicating a greater chance of success.

%Our aggregation methods consider statistical representativeness and relative importance within the overall plan structure, using factors such as graph distance or node degree.
% Our aggregation methods include minimum (min) and arithmetic average (mean), as well as weighted averages that consider the relative importance of a subtask within the plan structure, such as graph distance or node degree.
%
%(e.g., the ``success'' prediction probability as a score)
We begin with simple methods such as selecting the lowest score among all subtasks (\textit{min} aggregator), highlighting the weakest agent execution; and computing the arithmetic average of all scores (\textit{mean} aggregator), providing a balanced view of subtask performances.

We consider the relative importance of agents within the plan structure, as overall task performance depends on how effectively an agent's output transfers to others. To capture this, we propose two structural metrics for weighting agent scores. 
The distance-based metric emphasizes agents positioned closer to the source or sink nodes in the plan graph, under the assumption that these agents play a more critical role in the execution flow. This metric weights each agent's score inversely proportional to its distance from either the source or sink node (\textit{source distance} or \textit{sink distance} aggregator), ensuring that agents with greater structural influence have a higher impact on the aggregated score. Given the overall task \( T \) consisting of subtasks fulfilled by agents \( \{S_1, S_2, \ldots, S_m\} \), we denote the $i$-th agent $S_i$'s verifier score as \( \hat{y}_i \), which is predicted based on features extracted from its execution outputs. 

\begin{equation} 
\text{AggScore}_{\text{dist}}\left({\hat{y}_1, \hat{y}_2, \ldots, \hat{y}m}\right) = \frac{\sum{i=1}^{m} \frac{\hat{y}i}{d_i}}{\sum{i=1}^{m} \frac{1}{d_i}}, 
\end{equation}
where \( d_i \) denotes the shortest path distance from agent \( S_i \). 

Similarly, the degree-based metric assigns higher weights to agents with greater connectivity, reflecting their influence within the overall plan structure. It weights each agent's score based on the indegree or outdegree (\textit{indegree} or \textit{outdegree} aggregator) of its node:

\begin{equation} 
\text{AggScore}_{\text{deg}}\left({\hat{y}_1, \hat{y}_2, \ldots, \hat{y}m}\right) = \frac{\sum{i=1}^{m} \text{deg}_i \cdot \hat{y}i}{\sum{i=1}^{m} \text{deg}_i}, 
\end{equation}
% To determine whether an overall execution failure has been detected, we compare the aggregated result to a threshold parameter \( th \). An execution failure is flagged if \( \text{Aggregate}(\{\hat{y}_1, \hat{y}_2, \ldots, \hat{y}_m\}) > th \); otherwise, it is not.
where \( \hat{y}_i \) represents the prediction score for subtask \( S_i \), and \( \text{deg}_i \) denotes the degree (either indegree or outdegree) of subtask \( S_i \) in the plan DAG. 

We later assess the accuracy of the aggregated verification results by comparing the task’s gold label, indicating actual plan success, with the final agent label predicted by the verifier. This comparison evaluates how effectively the verifier is aggregated to predict task failure (\S~\ref{sec:agg_results}).

\begin{table*}[tp]
\caption{Human-designed agent registry for mathematical reasoning tasks} %\todo{check capitalization, content - something may missing in prev version with two datasets}}
\centering
\footnotesize 
\begin{tabular}{l p{4cm} p{2cm} p{4cm} p{3cm}}
\toprule
\textbf{Agent}& \textbf{Role} & \textbf{Input} & \textbf{Output} & \textbf{Output Format} \\ 
\midrule
\textbf{Identify Operands} & Identify operands with text description of each operands & Math question & List of operand names with their values & \{<name>: Number, ...\} \\ 
%\midrule
\textbf{Add} & Add numbers or dates & List of operands & One summed value & Number or Date \\ 
%\midrule
\textbf{Subtract} & Subtract numbers or dates & List of operands & One subtracted value & Number or Date \\ 
%\midrule
\textbf{Multiply} & Multiply numbers & List of operands & One multiplied value & Number \\ 
%\midrule
\textbf{Divide} & Divide numbers & List of operands & One divided value & Number \\ 
\textbf{Filter} & Filter a list based on a condition & List, condition & Filtered list & List \\
\textbf{Sort} & Sort a list by an attribute & List, attribute & Sorted list & List \\
\textbf{Convert Format} & Convert input from one format to another format & Text, format & Formatted text & Text \\
\textbf{Date Lookup} & Identify year, month, and day from a natural language description & Text & Date & Date \\
\bottomrule
\end{tabular}
\label{tab:agent_registry}
\end{table*}

\section{Case Study: Mathematical Reasoning}

We demonstrate the effectiveness of \name{} on mathematical reasoning tasks.
Math reasoning problems can be naturally decomposed into step-by-step plans, allowing us to focus on evaluating agent execution failures while ensuring that the results are easily verifiable by humans.

\subsection{Experiment Setting}

\subsubsection{Datasets}
We evaluate our pipeline on four math reasoning datasets: GSM8K~\cite{cobbe2021training} and Date Understanding (C3), Multi-Step Arithmetic (C11), Object Counting (C13) from BIG-Bench Hard (BBH)~\citep{suzgun2022challenging}.
The GSM8K dataset consists of grade-school-level math word problems written in natural language, whereas the Multi-Step Arithmetic dataset is presented in equation format and involves more complex problems that require multiple steps. Date Understanding focuses on reasoning about and performing operations with dates, while Object Counting involves enumerating objects of interests.

\subsubsection{Planning and Agent Execution}
For each task, our planning agent generates a DAG plan to solve it using a human-designed agent registry (detailed in Table~\ref{tab:agent_registry}).
We manually filter out instances where the generated plans are invalid,\footnote{We consider both structural validity (e.g., missing dependencies between subtasks) and semantic correctness (e.g., incorrect agent assignment or faulty instructions).} keeping only those with valid plans.
% \dz{Do we have the definition of plan validity somewhere}
The average number of subtasks per plan is 4 for GSM8K, 3.5 for Date Understanding, 2.5 for Object Counting, and 5.6 for Multistep Arithmetic.
% The remaining tasks result in an average of 2.5 (Object Counting) to 5.6 (Multistep Arithmetic) subtasks per plan.\yy{other two dataset stats}
Each subtask in a plan is then executed by an assigned agent.
Both the planning agent and agents in agent registry use GPT-4o with a temperature of 0.1. All prompts for planning, execution, and verification are provided in Appendix~\ref{app:prompts}.

\subsubsection{Agent Execution Verfication}
\paragraph{Gold (Execution Failure) Label Annotation}
% The goal of the verifier is to identify which execution agent fails to perform its assigned subtask according to human judgment.
To train the verifier, human annotators label agent's execution failures based on agent criteria, expected inputs and outputs, and input information (\S~\ref{sec:verifier_label}).
For GSM8K, we collected 1,975 subtasks from 497 tasks, each subtask labeled by three annotators from crowdsourcing (Appendix~\ref{app:user_study}). 
Due to low inter-rater reliability (Fleiss’ kappa: 0.41), only unanimously labeled samples (973 subtasks) were retained for training.
For the BBH datasets, the three authors handled annotations to ensure quality while reducing costs.

\paragraph{Feature Collection}
We extracted 26 execution features as described in \S~\ref{sec:verifier}. 
For self-consistency, we used the same model as the corresponding execution agent with a temperature of 0.7 to generate diverse outputs, following~\citet{xiong2023can}.
For external LLM judges, we employed GPT-4o with a temperature of 0.1.
%To evaluate the correctness of each step, we collect three types of signals from the agent's execution: uncertainty, self-consistency, and per-criteria scores provided by external LLM judges.

% \begin{figure}
%   \includegraphics[width=0.95\linewidth,trim={0 0.5cm 22.6cm 2.5cm},clip]{fig/subtask_verifier_each.pdf}
%   \caption{Verifier accuracy across datasets: average over 5-fold cross-validation}
%   \Description{\todo{add}}
%   \label{fig:sub_verifier_each}
% \end{figure}

\subsection{Verifier Results for Agent Failures}\label{sec:verifier_results}

\begin{figure}[t!]
  \includegraphics[width=0.95\linewidth]
  {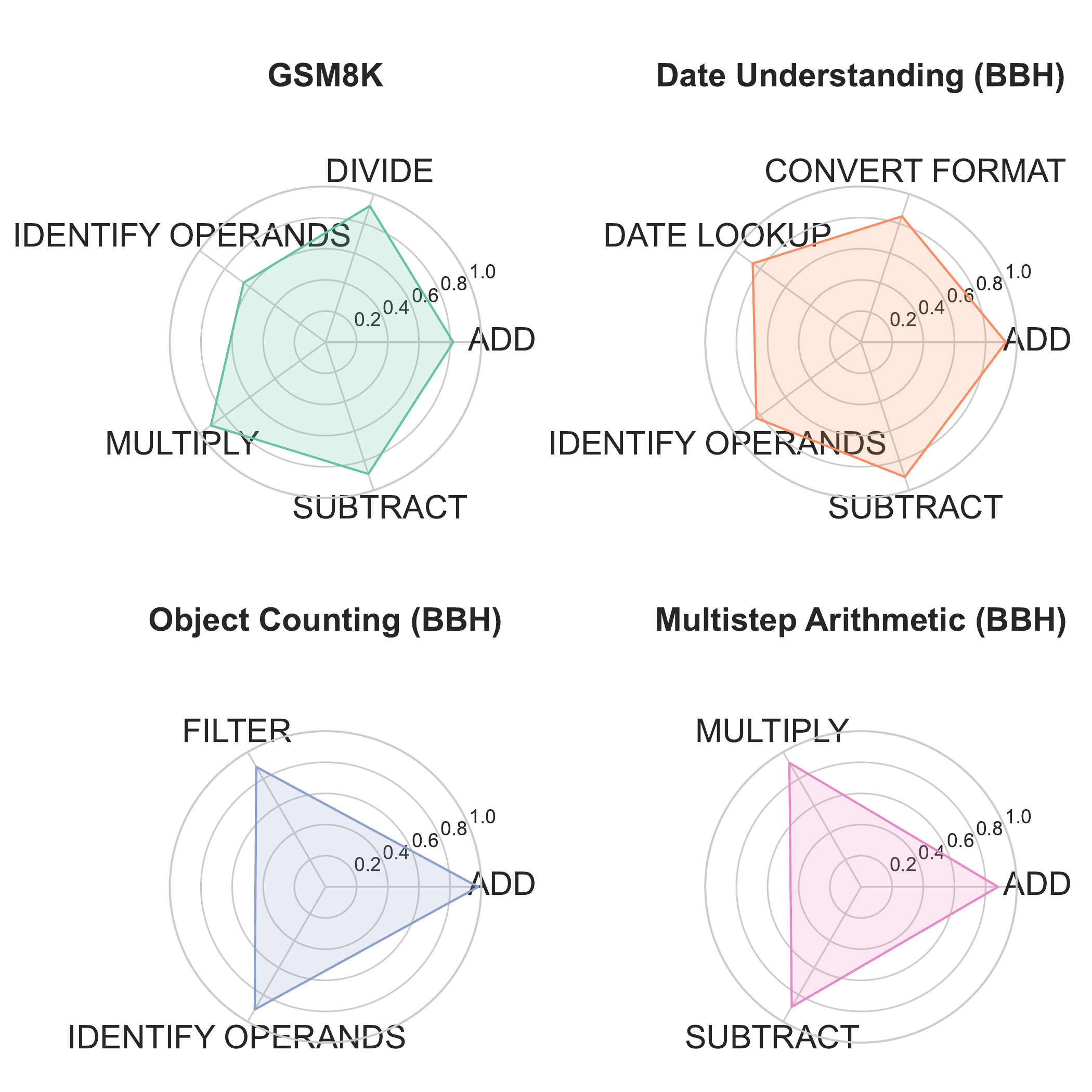}
  \caption{Verifier accuracy across datasets. The test accuracy remains consistently high across subtasks, without bias toward any specific one. Similar subtasks, like "Add" and "Subtract," which share the same criteria, also show comparable accuracies across all datasets.}
  \Description{\todo{add}}
  \label{fig:subtask_verifier_performance}
\end{figure}
 %closely followed by XGBoost (Fig.~\ref{fig:sub_verifier_each}).
 %app:different_models
Using features from all agent outputs in the plan, we train a simple machine learning model, Random Forest model, which achieve a high average accuracy (0.88\%) across four datasets.\footnote{We compared several ML models and select the one with the highest accuracy (Appendix~\ref{app:different_models}).}, suggesting the verifier's effectiveness in identifying failed agent executions.

%Our verifier is designed to function as a unified model across all agents. 
Next, we investigate whether its performance varies across different agents. 
As shown in Figure~\ref{fig:subtask_verifier_performance}, the test accuracy remains consistently high across various subtasks except ``identify operands'' in GSM8K where agents often struggle with accurate formatting. 
Moreover, similar subtasks---such as ``Add'' and ``Subtract'', which share the same subjective criteria---exhibit comparable test accuracies across all datasets, suggesting their generalizability to other tasks.

\begin{figure}[!t]
  \includegraphics[width=0.95\linewidth]
  {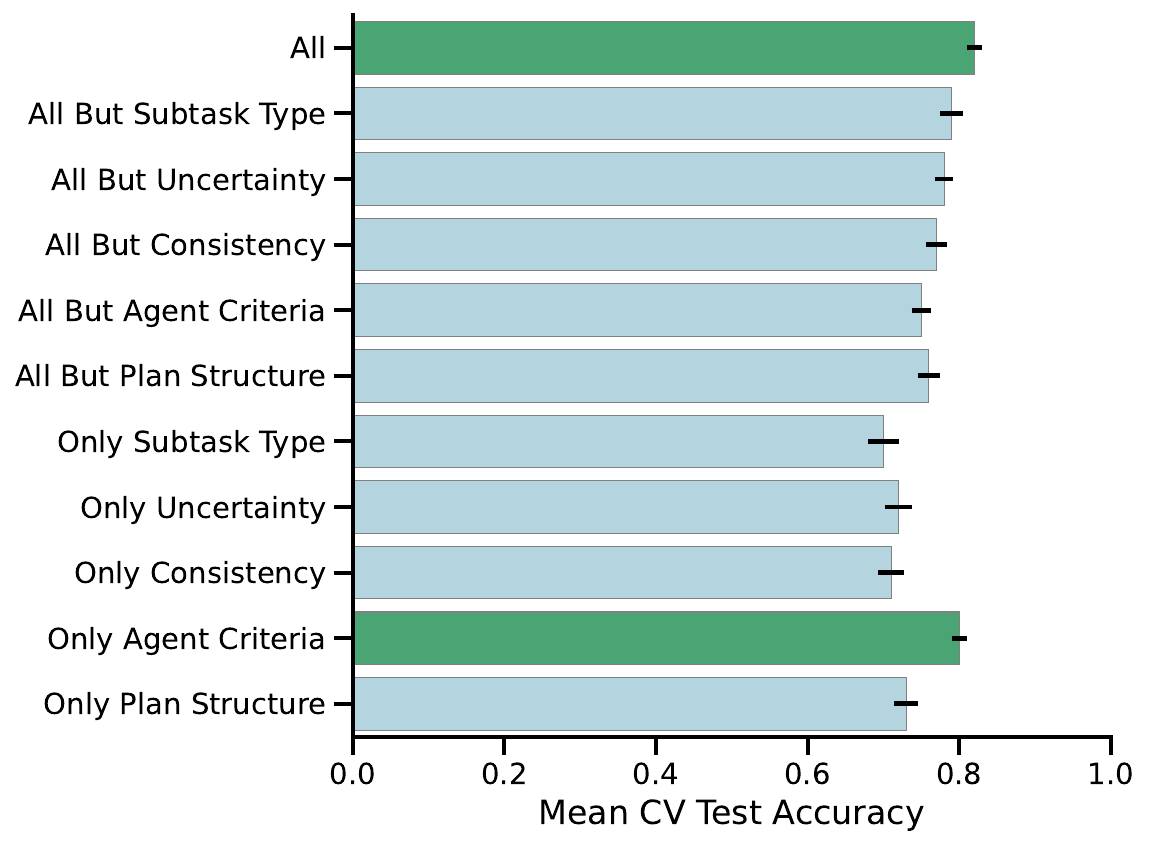}
  \caption{Ablation study on different feature configurations evaluating verifiers' test accuracy. Human-defined agent criteria feature enhances its performance, showing the highest accuracy when all features are used.}
  \Description{}
  \label{fig:ablation_study}
\end{figure}

To further analyze verifier behavior, we conduct an ablation study with different features to predict agent execution failures.
%\todo{check which dataset is this}, 
Our verifier achieves the highest performance with all features included while excluding any single feature leads to a drop in performance (Figure~\ref{fig:ablation_study}).
This suggests that the features provide complementary information, each playing a distinct role in model accuracy.
Notably, the agent criteria features has the largest performance drop when removed, suggesting that incorporating human-defined agent criteria from external LLM judges improves the verifiers' ability to align their predictions with human priorities.
On the other hand, the consistency-related features had minimal impact on performance, implying they may be less critical.%, particularly in scenarios where reducing prompting costs is a priority.

% \begin{table}[t]
% \centering
% \caption{Mean CV (Fold=5) Test accuracy analysis on feature ablation \todo{consistent feature group names across sections}}
% \begin{tabular}{l c}
% \toprule
% \textbf{Features}                                 & \textbf{Accuracy}               \\ \midrule
% \textbf{All Features}                             & \textbf{0.87 ± 0.03}                                  \\ \midrule
% \textbf{All But Subtask Features}                 & 0.86 ± 0.03                         \\ 
% \textbf{All But Uncertainty Features}             & 0.86 ± 0.03                      \\ 
% \textbf{All But Consistency Features}             & \textbf{0.87 ± 0.03}                         \\ 
% \textbf{All But Criteria Features}                & \textbf{0.84 ± 0.03}                         \\ 
% \textbf{All But Subtask Dependency Features}           & 0.86 ± 0.03                         \\ \midrule
% \textbf{Subtask Features}                         & 0.76 ± 0.02                         \\ 
% \textbf{Uncertainty Features}                     & 0.78 ± 0.03                         \\ 
% \textbf{Consistency Features}                     & 0.79 ± 0.02                         \\ 
% \textbf{Criteria Features}                        & \textbf{0.85 ± 0.03}                         \\ 
% \textbf{Subtask Dependency Features}                   & 0.75 ± 0.02                         \\ 
% \bottomrule
% \end{tabular}
% \label{tab:feature_ablation}
% \end{table}

\subsection{Aggregator Results for Task Failures}\label{sec:agg_results}
To predict whether the overall task is failing, we present a few aggregation metrics (aggregator), which combine subtask verification scores, that function as a \textit{task-level verifier}.\footnote{A task is considered successfully solved if the last step's execution output is linguistically equivalent~\citep{li2024panda} to the gold answer from the original dataset.}
By leveraging the aggregated score from the task verifier, users can prioritize and investigate tasks that are most likely to generate false LLM outputs. They can then analyze the reasons behind a task's failure by examining the flagged subtasks identified by the agent verifier.

\begin{figure}[thb]
  \includegraphics[width=0.95\linewidth]
  {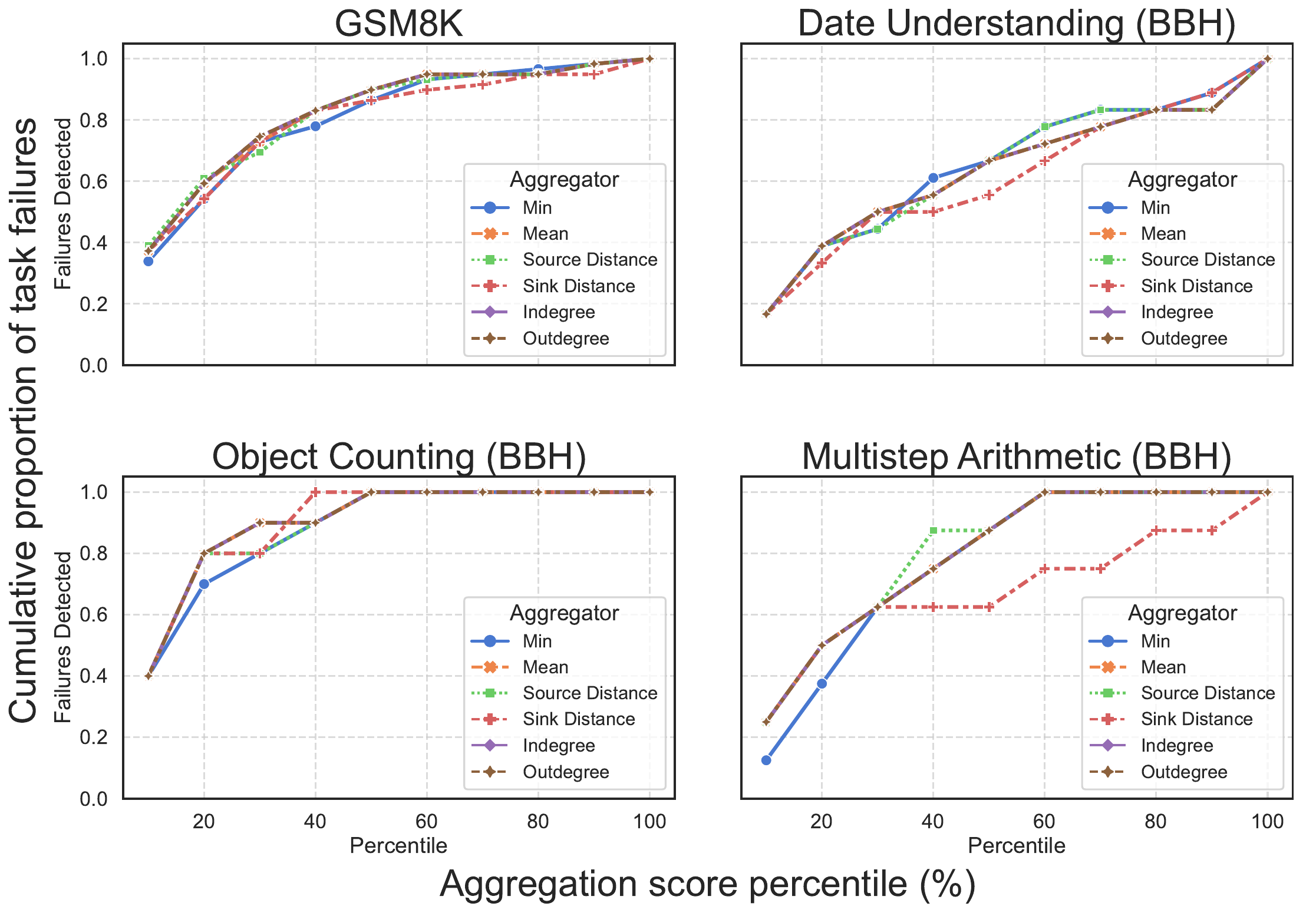}
  \caption{Aggregation performance measured by failure rate across aggregation score percentiles. They all show an upward trend, suggesting that they can help users prioritize tasks more likely to fail, when the labor budget is limited, allowing auditing of high-risk tasks first. Overall, \textit{mean} and \textit{outdegree} showed stable performance across datasets and can be used as default aggregation metrics for new datasets.}
  \Description{\todo{add}}
  \label{fig:aggregator_accuracy}
\end{figure}

Figure~\ref{fig:aggregator_accuracy} shows task-level verification performance from different aggregation methods, with lower score indicating a greater chance of failure. $x$-axis represents percentiles of ranked aggregation scores, and $y$-axis shows the cumulative ratio of detected failures within each percentile relative to all failed tasks. The curves closer to the top-left corner indicate better performance. \textit{Sink distance} aggregator identified all failures fastest in object counting but was slowest in date understanding. \textit{Source distance} aggregator generally outperformed \textit{sink distance}, except in GSM8K. This suggests that, for GSM8K, proximity to the starting nodes is a stronger indicator than proximity to the final node; GSM8K's typical first subtask---identifying operands---is a common source of error. Although no single aggregator consistently outperforms others across all datasets, they all show an upward trend, suggesting that they can help users prioritize tasks more likely to fail. This can be especially useful when labor is limited, allowing auditing of high-risk tasks first.  Overall, \textit{mean} and \textit{outdegree} showed stable performance across datasets and can be used as default aggregation metrics for new datasets.

% \begin{table}[t!]
% \centering
% \footnotesize
% \setlength{\tabcolsep}{3pt}
% \renewcommand{\arraystretch}{1.1}
% \caption{Best accuracy for each aggregator across datasets}
% \resizebox{\columnwidth}{!}{
% \begin{tabular}{lcccc}
% \toprule
% \textbf{Aggregator} & \textbf{GSM8K} & \makecell{\textbf{Date} \\ \textbf{Understanding}}  & \makecell{\textbf{Object} \\ \textbf{Counting}} & \makecell{\textbf{Multistep} \\ \textbf{Arithmetic}} \\
% \midrule
% Min & 0.825 & \textbf{0.714} & 0.925 & 0.774 \\
% Mean & 0.842 & 0.686 & 0.950 & 0.806 \\
% Source Distance & 0.833 & 0.686 & 0.950 & \textbf{0.839} \\
% Sink Distance & \textbf{0.871} & \textbf{0.714} & 0.950 & 0.806 \\
% Indegree & 0.863 & 0.686 & 0.950 & 0.806 \\
% Outdegree & 0.863 & 0.686 & 0.950 & 0.806 \\
% \bottomrule
% \end{tabular}
% }
% \label{tab:aggregator_accuracy}
% \end{table}

\section{Conclusion and Future Work}
In this work, we introduce \name{}, a human-centered evaluation framework that verifies agent execution failures; this encourages reliability and interpretability in humans using compound AI systems. By applying a verifier that assesses each agent's outputs through a combination of human-defined criteria, agent output uncertainty, and agent dependencies within the plan. Thus, \name{} also captures error propagation within the plan, facilitating better human-agent interaction. We also present a verifier-driven task failure metric that help users detect tasks prior to their auditing. Thus, \name{} enhances accountability on agent performance and labor efficiency by enabling granular human inspection of failing agents and the underlying reasons for their failures.
% For future work, we encourage applying our human-centered evaluation framework to open agentic systems which involves diverse execution agents with varying input formats and error tolerances. In these systems, capturing the dynamics of error propagation becomes even more challenging, highlighting the need for robust, human-centered evaluation methods for human-agent interaction. 
%\todo{add future work ideas, user study}

Future work remains for open agentic systems that involve diverse execution agents with more complexity and dynamic error propagation, highlighting the need for robust, human-centered evaluation methods for human-agent interaction.
For future work, we first plan to conduct a crowdsourced user study that evaluates \name{}'s subtask and task-level verification performance and usability, comparing outcomes with and without \name{}. Second, building on our aggregation metrics, we will develop an advanced aggregator that directly leverages user-centric aspects like free-form user feedback and a real-time incentive system. Finally, we will expand our framework to broader applications, such as open-domain question answering and fact-checking, to support a wider range of human-agent interactive systems. Thus, we aim to cultivate synergy between humans and LLM agents by designing compound AI systems that align with human needs, enabling collaboration on real-world tasks that require intricate problem-solving.
% Thus, we AI practitioners to attain actionable feedback while auditing agents' outputs, and humans to refine and improve agent performance in various real-world applications.

%multiple evaluation aspects would enhance adaptability and effectiveness across diverse applications. we plan to explore more sophisticated aggregations. The aggregation schemes tested in this work focus on single aspects, each exhibiting different strengths depending on the dataset. 

% \begin{acks}
% \end{acks}

\bibliographystyle{ACM-Reference-Format}
% \balance
\bibliography{custom,dz}


\begin{thebibliography}{42}

%%% the Web address.

\ifx \showCODEN    \undefined \def \showCODEN     #1{\unskip}     \fi
\ifx \showISBNx    \undefined \def \showISBNx     #1{\unskip}     \fi
\ifx \showISBNxiii \undefined \def \showISBNxiii  #1{\unskip}     \fi
\ifx \showISSN     \undefined \def \showISSN      #1{\unskip}     \fi
\ifx \showLCCN     \undefined \def \showLCCN      #1{\unskip}     \fi
\ifx \shownote     \undefined \def \shownote      #1{#1}          \fi
\ifx \showarticletitle \undefined \def \showarticletitle #1{#1}   \fi
\ifx \showURL      \undefined \def \showURL       {\relax}        \fi
% invisible to TeX
\providecommand\bibfield[2]{#2}
\providecommand\bibinfo[2]{#2}
\providecommand\natexlab[1]{#1}
\providecommand\showeprint[2][]{arXiv:#2}

\bibitem[Arabzadeh et~al\mbox{.}(2024)]%
        {arabzadeh2024towards}
\bibfield{author}{\bibinfo{person}{Negar Arabzadeh}, \bibinfo{person}{Julia Kiseleva}, \bibinfo{person}{Qingyun Wu}, \bibinfo{person}{Chi Wang}, \bibinfo{person}{Ahmed Awadallah}, \bibinfo{person}{Victor Dibia}, \bibinfo{person}{Adam Fourney}, {and} \bibinfo{person}{Charles Clarke}.} \bibinfo{year}{2024}\natexlab{}.
\newblock \showarticletitle{Towards better Human-Agent Alignment: Assessing Task Utility in LLM-Powered Applications}.
\newblock \bibinfo{journal}{\emph{arXiv preprint arXiv:2402.09015}} (\bibinfo{year}{2024}).
\newblock


\bibitem[Arawjo et~al\mbox{.}(2024)]%
        {arawjo2024chainforge}
\bibfield{author}{\bibinfo{person}{Ian Arawjo}, \bibinfo{person}{Chelse Swoopes}, \bibinfo{person}{Priyan Vaithilingam}, \bibinfo{person}{Martin Wattenberg}, {and} \bibinfo{person}{Elena~L. Glassman}.} \bibinfo{year}{2024}\natexlab{}.
\newblock \showarticletitle{ChainForge: A Visual Toolkit for Prompt Engineering and LLM Hypothesis Testing}. In \bibinfo{booktitle}{\emph{Proceedings of the CHI Conference on Human Factors in Computing Systems}} (Honolulu, HI, USA) \emph{(\bibinfo{series}{CHI '24})}. \bibinfo{publisher}{Association for Computing Machinery}, \bibinfo{address}{New York, NY, USA}, Article \bibinfo{articleno}{304}, \bibinfo{numpages}{18}~pages.
\newblock
\showISBNx{9798400703300}
\href{https://doi.org/10.1145/3613904.3642016}{doi:\nolinkurl{10.1145/3613904.3642016}}


\bibitem[Bottou(2010)]%
        {bottou2010large}
\bibfield{author}{\bibinfo{person}{L{\'e}on Bottou}.} \bibinfo{year}{2010}\natexlab{}.
\newblock \showarticletitle{Large-scale machine learning with stochastic gradient descent}. In \bibinfo{booktitle}{\emph{Proceedings of the 19th International Conference on Computational Statistics (COMPSTAT 2010)}}. \bibinfo{publisher}{Physica-Verlag HD}, \bibinfo{pages}{177--186}.
\newblock


\bibitem[Breiman et~al\mbox{.}(1986)]%
        {breiman1986classification}
\bibfield{author}{\bibinfo{person}{Leo Breiman}, \bibinfo{person}{Jerome Friedman}, \bibinfo{person}{Richard~A Olshen}, {and} \bibinfo{person}{Charles~J Stone}.} \bibinfo{year}{1986}\natexlab{}.
\newblock \bibinfo{booktitle}{\emph{Classification and Regression Trees}}.
\newblock \bibinfo{publisher}{Wadsworth}.
\newblock


\bibitem[Chen and Mueller(2024)]%
        {chen2024quantifying}
\bibfield{author}{\bibinfo{person}{Jiuhai Chen} {and} \bibinfo{person}{Jonas Mueller}.} \bibinfo{year}{2024}\natexlab{}.
\newblock \showarticletitle{Quantifying uncertainty in answers from any language model and enhancing their trustworthiness}. In \bibinfo{booktitle}{\emph{Proceedings of the 62nd Annual Meeting of the Association for Computational Linguistics (Volume 1: Long Papers)}}. \bibinfo{pages}{5186--5200}.
\newblock


\bibitem[Cheng et~al\mbox{.}(2024)]%
        {cheng2024exploring}
\bibfield{author}{\bibinfo{person}{Yuheng Cheng}, \bibinfo{person}{Ceyao Zhang}, \bibinfo{person}{Zhengwen Zhang}, \bibinfo{person}{Xiangrui Meng}, \bibinfo{person}{Sirui Hong}, \bibinfo{person}{Wenhao Li}, \bibinfo{person}{Zihao Wang}, \bibinfo{person}{Zekai Wang}, \bibinfo{person}{Feng Yin}, \bibinfo{person}{Junhua Zhao}, {et~al\mbox{.}}} \bibinfo{year}{2024}\natexlab{}.
\newblock \showarticletitle{Exploring Large Language Model based Intelligent Agents: Definitions, Methods, and Prospects}.
\newblock \bibinfo{journal}{\emph{CoRR}} (\bibinfo{year}{2024}).
\newblock


\bibitem[Cobbe et~al\mbox{.}(2021)]%
        {cobbe2021training}
\bibfield{author}{\bibinfo{person}{Karl Cobbe}, \bibinfo{person}{Vineet Kosaraju}, \bibinfo{person}{Mohammad Bavarian}, \bibinfo{person}{Mark Chen}, \bibinfo{person}{Heewoo Jun}, \bibinfo{person}{Lukasz Kaiser}, \bibinfo{person}{Matthias Plappert}, \bibinfo{person}{Jerry Tworek}, \bibinfo{person}{Jacob Hilton}, \bibinfo{person}{Reiichiro Nakano}, {et~al\mbox{.}}} \bibinfo{year}{2021}\natexlab{}.
\newblock \showarticletitle{Training verifiers to solve math word problems}.
\newblock \bibinfo{journal}{\emph{arXiv preprint arXiv:2110.14168}} (\bibinfo{year}{2021}).
\newblock


\bibitem[Cox(1958)]%
        {cox1958regression}
\bibfield{author}{\bibinfo{person}{David~R Cox}.} \bibinfo{year}{1958}\natexlab{}.
\newblock \showarticletitle{The regression analysis of binary sequences}.
\newblock \bibinfo{journal}{\emph{Journal of the Royal Statistical Society. Series B (Methodological)}} \bibinfo{volume}{20}, \bibinfo{number}{2} (\bibinfo{year}{1958}), \bibinfo{pages}{215--242}.
\newblock


\bibitem[Freund and Schapire(1997)]%
        {freund1997decision}
\bibfield{author}{\bibinfo{person}{Yoav Freund} {and} \bibinfo{person}{Robert~E Schapire}.} \bibinfo{year}{1997}\natexlab{}.
\newblock \showarticletitle{A decision-theoretic generalization of on-line learning and an application to boosting}.
\newblock \bibinfo{journal}{\emph{J. Comput. System Sci.}} \bibinfo{volume}{55}, \bibinfo{number}{1} (\bibinfo{year}{1997}), \bibinfo{pages}{119--139}.
\newblock


\bibitem[Geng et~al\mbox{.}(2024)]%
        {geng-etal-2024-survey}
\bibfield{author}{\bibinfo{person}{Jiahui Geng}, \bibinfo{person}{Fengyu Cai}, \bibinfo{person}{Yuxia Wang}, \bibinfo{person}{Heinz Koeppl}, \bibinfo{person}{Preslav Nakov}, {and} \bibinfo{person}{Iryna Gurevych}.} \bibinfo{year}{2024}\natexlab{}.
\newblock \showarticletitle{A Survey of Confidence Estimation and Calibration in Large Language Models}. In \bibinfo{booktitle}{\emph{Proceedings of the 2024 Conference of the North American Chapter of the Association for Computational Linguistics: Human Language Technologies (Volume 1: Long Papers)}}, \bibfield{editor}{\bibinfo{person}{Kevin Duh}, \bibinfo{person}{Helena Gomez}, {and} \bibinfo{person}{Steven Bethard}} (Eds.). \bibinfo{publisher}{Association for Computational Linguistics}, \bibinfo{address}{Mexico City, Mexico}, \bibinfo{pages}{6577--6595}.
\newblock
\href{https://doi.org/10.18653/v1/2024.naacl-long.366}{doi:\nolinkurl{10.18653/v1/2024.naacl-long.366}}


\bibitem[Ghallab et~al\mbox{.}(2004)]%
        {ghallab2004automated}
\bibfield{author}{\bibinfo{person}{Malik Ghallab}, \bibinfo{person}{Dana Nau}, {and} \bibinfo{person}{Paolo Traverso}.} \bibinfo{year}{2004}\natexlab{}.
\newblock \bibinfo{booktitle}{\emph{Automated Planning: theory and practice}}.
\newblock \bibinfo{publisher}{Elsevier}.
\newblock


\bibitem[Goyal et~al\mbox{.}(2024)]%
        {goyal2024designing}
\bibfield{author}{\bibinfo{person}{Nitesh Goyal}, \bibinfo{person}{Minsuk Chang}, {and} \bibinfo{person}{Michael Terry}.} \bibinfo{year}{2024}\natexlab{}.
\newblock \showarticletitle{Designing for Human-Agent Alignment: Understanding what humans want from their agents}. In \bibinfo{booktitle}{\emph{Extended Abstracts of the CHI Conference on Human Factors in Computing Systems}}. \bibinfo{pages}{1--6}.
\newblock


\bibitem[Grunde-McLaughlin et~al\mbox{.}(2024)]%
        {grundemclaughlin2024designingllmchainsadapting}
\bibfield{author}{\bibinfo{person}{Madeleine Grunde-McLaughlin}, \bibinfo{person}{Michelle~S. Lam}, \bibinfo{person}{Ranjay Krishna}, \bibinfo{person}{Daniel~S. Weld}, {and} \bibinfo{person}{Jeffrey Heer}.} \bibinfo{year}{2024}\natexlab{}.
\newblock \bibinfo{title}{Designing LLM Chains by Adapting Techniques from Crowdsourcing Workflows}.
\newblock
\showeprint[arxiv]{2312.11681}~[cs.HC]
\urldef\tempurl%
\url{https://arxiv.org/abs/2312.11681}
\showURL{%
\tempurl}


\bibitem[Huang et~al\mbox{.}(2024)]%
        {huang2024understanding}
\bibfield{author}{\bibinfo{person}{Xu Huang}, \bibinfo{person}{Weiwen Liu}, \bibinfo{person}{Xiaolong Chen}, \bibinfo{person}{Xingmei Wang}, \bibinfo{person}{Hao Wang}, \bibinfo{person}{Defu Lian}, \bibinfo{person}{Yasheng Wang}, \bibinfo{person}{Ruiming Tang}, {and} \bibinfo{person}{Enhong Chen}.} \bibinfo{year}{2024}\natexlab{}.
\newblock \showarticletitle{Understanding the planning of LLM agents: A survey}.
\newblock \bibinfo{journal}{\emph{arXiv preprint arXiv:2402.02716}} (\bibinfo{year}{2024}).
\newblock


\bibitem[Huang et~al\mbox{.}(2023)]%
        {huang2023lookleapexploratorystudy}
\bibfield{author}{\bibinfo{person}{Yuheng Huang}, \bibinfo{person}{Jiayang Song}, \bibinfo{person}{Zhijie Wang}, \bibinfo{person}{Shengming Zhao}, \bibinfo{person}{Huaming Chen}, \bibinfo{person}{Felix Juefei-Xu}, {and} \bibinfo{person}{Lei Ma}.} \bibinfo{year}{2023}\natexlab{}.
\newblock \bibinfo{title}{Look Before You Leap: An Exploratory Study of Uncertainty Measurement for Large Language Models}.
\newblock
\showeprint[arxiv]{2307.10236}~[cs.SE]
\urldef\tempurl%
\url{https://arxiv.org/abs/2307.10236}
\showURL{%
\tempurl}


\bibitem[Ibrahim et~al\mbox{.}(2024)]%
        {ibrahim2024beyond}
\bibfield{author}{\bibinfo{person}{Lujain Ibrahim}, \bibinfo{person}{Saffron Huang}, \bibinfo{person}{Lama Ahmad}, {and} \bibinfo{person}{Markus Anderljung}.} \bibinfo{year}{2024}\natexlab{}.
\newblock \showarticletitle{Beyond static AI evaluations: advancing human interaction evaluations for LLM harms and risks}.
\newblock \bibinfo{journal}{\emph{arXiv preprint arXiv:2405.10632}} (\bibinfo{year}{2024}).
\newblock


\bibitem[Jaeger et~al\mbox{.}(2013)]%
        {jaeger2013uncertainty}
\bibfield{author}{\bibinfo{person}{Laure Jaeger}, \bibinfo{person}{Tom Jorquera}, \bibinfo{person}{Sylvain Lemouzy}, \bibinfo{person}{Christian Gogu}, \bibinfo{person}{St{\'e}phane Segonds}, {and} \bibinfo{person}{Christian Bes}.} \bibinfo{year}{2013}\natexlab{}.
\newblock \showarticletitle{Uncertainty propagation in multi-agent systems for multidisciplinary optimization problems}. In \bibinfo{booktitle}{\emph{10th World Congress on Structural and Multidisciplinary Optimization (WCSMO 10)}}. \bibinfo{pages}{pp--1}.
\newblock


\bibitem[John and Langley(1995)]%
        {john1995estimating}
\bibfield{author}{\bibinfo{person}{Geoffrey~H John} {and} \bibinfo{person}{Pat Langley}.} \bibinfo{year}{1995}\natexlab{}.
\newblock \showarticletitle{Estimating continuous distributions in Bayesian classifiers}. In \bibinfo{booktitle}{\emph{Proceedings of the 11th Conference on Uncertainty in Artificial Intelligence (UAI 1995)}}. \bibinfo{publisher}{Morgan Kaufmann}, \bibinfo{pages}{338--345}.
\newblock


\bibitem[Kim et~al\mbox{.}(2024)]%
        {kim2024evallm}
\bibfield{author}{\bibinfo{person}{Tae~Soo Kim}, \bibinfo{person}{Yoonjoo Lee}, \bibinfo{person}{Jamin Shin}, \bibinfo{person}{Young-Ho Kim}, {and} \bibinfo{person}{Juho Kim}.} \bibinfo{year}{2024}\natexlab{}.
\newblock \showarticletitle{EvalLM: Interactive Evaluation of Large Language Model Prompts on User-Defined Criteria}. In \bibinfo{booktitle}{\emph{Proceedings of the CHI Conference on Human Factors in Computing Systems}} (Honolulu, HI, USA) \emph{(\bibinfo{series}{CHI '24})}. \bibinfo{publisher}{Association for Computing Machinery}, \bibinfo{address}{New York, NY, USA}, Article \bibinfo{articleno}{306}, \bibinfo{numpages}{21}~pages.
\newblock
\showISBNx{9798400703300}
\href{https://doi.org/10.1145/3613904.3642216}{doi:\nolinkurl{10.1145/3613904.3642216}}


\bibitem[LangChain(2013)]%
        {langchain}
\bibfield{author}{\bibinfo{person}{LangChain}.} \bibinfo{year}{2013}\natexlab{}.
\newblock \bibinfo{title}{LangChain}.
\newblock \bibinfo{howpublished}{\url{https://github.com/langchain-ai/langchain}}.
\newblock


\bibitem[Li et~al\mbox{.}(2024)]%
        {li2024panda}
\bibfield{author}{\bibinfo{person}{Zongxia Li}, \bibinfo{person}{Ishani Mondal}, \bibinfo{person}{Yijun Liang}, \bibinfo{person}{Huy Nghiem}, {and} \bibinfo{person}{Jordan~Lee Boyd-Graber}.} \bibinfo{year}{2024}\natexlab{}.
\newblock \showarticletitle{PANDA (Pedantic ANswer-correctness Determination and Adjudication): Improving Automatic Evaluation for Question Answering and Text Generation}.
\newblock \bibinfo{journal}{\emph{arXiv preprint arXiv:2402.11161}} (\bibinfo{year}{2024}).
\newblock


\bibitem[Liao and Xiao(2023)]%
        {liao2023rethinking}
\bibfield{author}{\bibinfo{person}{Q~Vera Liao} {and} \bibinfo{person}{Ziang Xiao}.} \bibinfo{year}{2023}\natexlab{}.
\newblock \showarticletitle{Rethinking model evaluation as narrowing the socio-technical gap}.
\newblock \bibinfo{journal}{\emph{arXiv preprint arXiv:2306.03100}} (\bibinfo{year}{2023}).
\newblock


\bibitem[Lin and Martelaro(2024)]%
        {jigsaw}
\bibfield{author}{\bibinfo{person}{David Chuan-En Lin} {and} \bibinfo{person}{Nikolas Martelaro}.} \bibinfo{year}{2024}\natexlab{}.
\newblock \showarticletitle{Jigsaw: Supporting Designers to Prototype Multimodal Applications by Chaining AI Foundation Models}. In \bibinfo{booktitle}{\emph{Proceedings of the CHI Conference on Human Factors in Computing Systems}} (Honolulu, HI, USA) \emph{(\bibinfo{series}{CHI '24})}. \bibinfo{publisher}{Association for Computing Machinery}, \bibinfo{address}{New York, NY, USA}, Article \bibinfo{articleno}{4}, \bibinfo{numpages}{15}~pages.
\newblock
\showISBNx{9798400703300}
\href{https://doi.org/10.1145/3613904.3641920}{doi:\nolinkurl{10.1145/3613904.3641920}}


\bibitem[Lou et~al\mbox{.}(2024)]%
        {lou2024largelanguagemodelinstruction}
\bibfield{author}{\bibinfo{person}{Renze Lou}, \bibinfo{person}{Kai Zhang}, {and} \bibinfo{person}{Wenpeng Yin}.} \bibinfo{year}{2024}\natexlab{}.
\newblock \bibinfo{title}{Large Language Model Instruction Following: A Survey of Progresses and Challenges}.
\newblock
\showeprint[arxiv]{2303.10475}~[cs.CL]
\urldef\tempurl%
\url{https://arxiv.org/abs/2303.10475}
\showURL{%
\tempurl}


\bibitem[Madaan et~al\mbox{.}(2024)]%
        {madaan2024self}
\bibfield{author}{\bibinfo{person}{Aman Madaan}, \bibinfo{person}{Niket Tandon}, \bibinfo{person}{Prakhar Gupta}, \bibinfo{person}{Skyler Hallinan}, \bibinfo{person}{Luyu Gao}, \bibinfo{person}{Sarah Wiegreffe}, \bibinfo{person}{Uri Alon}, \bibinfo{person}{Nouha Dziri}, \bibinfo{person}{Shrimai Prabhumoye}, \bibinfo{person}{Yiming Yang}, {et~al\mbox{.}}} \bibinfo{year}{2024}\natexlab{}.
\newblock \showarticletitle{Self-refine: Iterative refinement with self-feedback}.
\newblock \bibinfo{journal}{\emph{Advances in Neural Information Processing Systems}}  \bibinfo{volume}{36} (\bibinfo{year}{2024}).
\newblock


\bibitem[Parmar et~al\mbox{.}(2019)]%
        {parmar2019review}
\bibfield{author}{\bibinfo{person}{Aakash Parmar}, \bibinfo{person}{Rakesh Katariya}, {and} \bibinfo{person}{Vatsal Patel}.} \bibinfo{year}{2019}\natexlab{}.
\newblock \showarticletitle{A review on random forest: An ensemble classifier}. In \bibinfo{booktitle}{\emph{International conference on intelligent data communication technologies and internet of things (ICICI) 2018}}. Springer, \bibinfo{pages}{758--763}.
\newblock


\bibitem[Prasad et~al\mbox{.}(2023)]%
        {prasad2023adapt}
\bibfield{author}{\bibinfo{person}{Archiki Prasad}, \bibinfo{person}{Alexander Koller}, \bibinfo{person}{Mareike Hartmann}, \bibinfo{person}{Peter Clark}, \bibinfo{person}{Ashish Sabharwal}, \bibinfo{person}{Mohit Bansal}, {and} \bibinfo{person}{Tushar Khot}.} \bibinfo{year}{2023}\natexlab{}.
\newblock \showarticletitle{Adapt: As-needed decomposition and planning with language models}.
\newblock \bibinfo{journal}{\emph{arXiv preprint arXiv:2311.05772}} (\bibinfo{year}{2023}).
\newblock


\bibitem[Rumelhart et~al\mbox{.}(1986)]%
        {rumelhart1986learning}
\bibfield{author}{\bibinfo{person}{David~E Rumelhart}, \bibinfo{person}{Geoffrey~E Hinton}, {and} \bibinfo{person}{Ronald~J Williams}.} \bibinfo{year}{1986}\natexlab{}.
\newblock \showarticletitle{Learning representations by back-propagating errors}.
\newblock \bibinfo{journal}{\emph{Nature}} \bibinfo{volume}{323}, \bibinfo{number}{6088} (\bibinfo{year}{1986}), \bibinfo{pages}{533--536}.
\newblock


\bibitem[Schapire(2013)]%
        {Schapire2013}
\bibfield{author}{\bibinfo{person}{Robert~E. Schapire}.} \bibinfo{year}{2013}\natexlab{}.
\newblock \bibinfo{booktitle}{\emph{Explaining AdaBoost}}.
\newblock \bibinfo{publisher}{Springer Berlin Heidelberg}, \bibinfo{address}{Berlin, Heidelberg}, \bibinfo{pages}{37--52}.
\newblock
\showISBNx{978-3-642-41136-6}
\href{https://doi.org/10.1007/978-3-642-41136-6_5}{doi:\nolinkurl{10.1007/978-3-642-41136-6_5}}


\bibitem[Shankar et~al\mbox{.}(2024)]%
        {shankar2024validates}
\bibfield{author}{\bibinfo{person}{Shreya Shankar}, \bibinfo{person}{JD Zamfirescu-Pereira}, \bibinfo{person}{Bj{\"o}rn Hartmann}, \bibinfo{person}{Aditya~G Parameswaran}, {and} \bibinfo{person}{Ian Arawjo}.} \bibinfo{year}{2024}\natexlab{}.
\newblock \showarticletitle{Who Validates the Validators? Aligning LLM-Assisted Evaluation of LLM Outputs with Human Preferences}.
\newblock \bibinfo{journal}{\emph{arXiv preprint arXiv:2404.12272}} (\bibinfo{year}{2024}).
\newblock


\bibitem[Stechly et~al\mbox{.}(2024)]%
        {stechly2024self}
\bibfield{author}{\bibinfo{person}{Kaya Stechly}, \bibinfo{person}{Karthik Valmeekam}, {and} \bibinfo{person}{Subbarao Kambhampati}.} \bibinfo{year}{2024}\natexlab{}.
\newblock \showarticletitle{On the Self-Verification Limitations of Large Language Models on Reasoning and Planning Tasks}.
\newblock \bibinfo{journal}{\emph{arXiv e-prints}} (\bibinfo{year}{2024}), \bibinfo{pages}{arXiv--2402}.
\newblock


\bibitem[Sumers et~al\mbox{.}(2023)]%
        {sumers2023cognitive}
\bibfield{author}{\bibinfo{person}{Theodore Sumers}, \bibinfo{person}{Shunyu Yao}, \bibinfo{person}{Karthik Narasimhan}, {and} \bibinfo{person}{Thomas Griffiths}.} \bibinfo{year}{2023}\natexlab{}.
\newblock \showarticletitle{Cognitive architectures for language agents}.
\newblock \bibinfo{journal}{\emph{Transactions on Machine Learning Research}} (\bibinfo{year}{2023}).
\newblock


\bibitem[Sun et~al\mbox{.}(2023)]%
        {sun2023adaplanner}
\bibfield{author}{\bibinfo{person}{Haotian Sun}, \bibinfo{person}{Yuchen Zhuang}, \bibinfo{person}{Lingkai Kong}, \bibinfo{person}{Bo Dai}, {and} \bibinfo{person}{Chao Zhang}.} \bibinfo{year}{2023}\natexlab{}.
\newblock \showarticletitle{AdaPlanner: Adaptive Planning from Feedback with Language Models}. In \bibinfo{booktitle}{\emph{Thirty-seventh Conference on Neural Information Processing Systems}}.
\newblock
\urldef\tempurl%
\url{https://openreview.net/forum?id=rnKgbKmelt}
\showURL{%
\tempurl}


\bibitem[Suzgun et~al\mbox{.}(2022)]%
        {suzgun2022challenging}
\bibfield{author}{\bibinfo{person}{Mirac Suzgun}, \bibinfo{person}{Nathan Scales}, \bibinfo{person}{Nathanael Sch{\"a}rli}, \bibinfo{person}{Sebastian Gehrmann}, \bibinfo{person}{Yi Tay}, \bibinfo{person}{Hyung~Won Chung}, \bibinfo{person}{Aakanksha Chowdhery}, \bibinfo{person}{Quoc~V Le}, \bibinfo{person}{Ed~H Chi}, \bibinfo{person}{Denny Zhou}, {et~al\mbox{.}}} \bibinfo{year}{2022}\natexlab{}.
\newblock \showarticletitle{Challenging big-bench tasks and whether chain-of-thought can solve them}.
\newblock \bibinfo{journal}{\emph{arXiv preprint arXiv:2210.09261}} (\bibinfo{year}{2022}).
\newblock


\bibitem[Wang et~al\mbox{.}(2024)]%
        {wang2024survey}
\bibfield{author}{\bibinfo{person}{Lei Wang}, \bibinfo{person}{Chen Ma}, \bibinfo{person}{Xueyang Feng}, \bibinfo{person}{Zeyu Zhang}, \bibinfo{person}{Hao Yang}, \bibinfo{person}{Jingsen Zhang}, \bibinfo{person}{Zhiyuan Chen}, \bibinfo{person}{Jiakai Tang}, \bibinfo{person}{Xu Chen}, \bibinfo{person}{Yankai Lin}, {et~al\mbox{.}}} \bibinfo{year}{2024}\natexlab{}.
\newblock \showarticletitle{A survey on large language model based autonomous agents}.
\newblock \bibinfo{journal}{\emph{Frontiers of Computer Science}} \bibinfo{volume}{18}, \bibinfo{number}{6} (\bibinfo{year}{2024}), \bibinfo{pages}{186345}.
\newblock


\bibitem[Wang et~al\mbox{.}(2022)]%
        {wang2022self}
\bibfield{author}{\bibinfo{person}{Xuezhi Wang}, \bibinfo{person}{Jason Wei}, \bibinfo{person}{Dale Schuurmans}, \bibinfo{person}{Quoc Le}, \bibinfo{person}{Ed Chi}, \bibinfo{person}{Sharan Narang}, \bibinfo{person}{Aakanksha Chowdhery}, {and} \bibinfo{person}{Denny Zhou}.} \bibinfo{year}{2022}\natexlab{}.
\newblock \showarticletitle{Self-consistency improves chain of thought reasoning in language models}.
\newblock \bibinfo{journal}{\emph{arXiv preprint arXiv:2203.11171}} (\bibinfo{year}{2022}).
\newblock


\bibitem[Wu et~al\mbox{.}(2022)]%
        {wu2022aichains}
\bibfield{author}{\bibinfo{person}{Tongshuang Wu}, \bibinfo{person}{Michael Terry}, {and} \bibinfo{person}{Carrie~Jun Cai}.} \bibinfo{year}{2022}\natexlab{}.
\newblock \showarticletitle{AI Chains: Transparent and Controllable Human-AI Interaction by Chaining Large Language Model Prompts}. In \bibinfo{booktitle}{\emph{Proceedings of the 2022 CHI Conference on Human Factors in Computing Systems}} (New Orleans, LA, USA) \emph{(\bibinfo{series}{CHI '22})}. \bibinfo{publisher}{Association for Computing Machinery}, \bibinfo{address}{New York, NY, USA}, Article \bibinfo{articleno}{385}, \bibinfo{numpages}{22}~pages.
\newblock
\showISBNx{9781450391573}
\href{https://doi.org/10.1145/3491102.3517582}{doi:\nolinkurl{10.1145/3491102.3517582}}


\bibitem[Xi et~al\mbox{.}(2025)]%
        {xi2025rise}
\bibfield{author}{\bibinfo{person}{Zhiheng Xi}, \bibinfo{person}{Wenxiang Chen}, \bibinfo{person}{Xin Guo}, \bibinfo{person}{Wei He}, \bibinfo{person}{Yiwen Ding}, \bibinfo{person}{Boyang Hong}, \bibinfo{person}{Ming Zhang}, \bibinfo{person}{Junzhe Wang}, \bibinfo{person}{Senjie Jin}, \bibinfo{person}{Enyu Zhou}, {et~al\mbox{.}}} \bibinfo{year}{2025}\natexlab{}.
\newblock \showarticletitle{The rise and potential of large language model based agents: A survey}.
\newblock \bibinfo{journal}{\emph{Science China Information Sciences}} \bibinfo{volume}{68}, \bibinfo{number}{2} (\bibinfo{year}{2025}), \bibinfo{pages}{121101}.
\newblock


\bibitem[Xiong et~al\mbox{.}(2024)]%
        {xiong2023can}
\bibfield{author}{\bibinfo{person}{Miao Xiong}, \bibinfo{person}{Zhiyuan Hu}, \bibinfo{person}{Xinyang Lu}, \bibinfo{person}{YIFEI LI}, \bibinfo{person}{Jie Fu}, \bibinfo{person}{Junxian He}, {and} \bibinfo{person}{Bryan Hooi}.} \bibinfo{year}{2024}\natexlab{}.
\newblock \showarticletitle{Can {LLM}s Express Their Uncertainty? An Empirical Evaluation of Confidence Elicitation in {LLM}s}. In \bibinfo{booktitle}{\emph{The Twelfth International Conference on Learning Representations}}.
\newblock
\urldef\tempurl%
\url{https://openreview.net/forum?id=gjeQKFxFpZ}
\showURL{%
\tempurl}


\bibitem[Yona et~al\mbox{.}(2024)]%
        {yona2024largelanguagemodelsfaithfully}
\bibfield{author}{\bibinfo{person}{Gal Yona}, \bibinfo{person}{Roee Aharoni}, {and} \bibinfo{person}{Mor Geva}.} \bibinfo{year}{2024}\natexlab{}.
\newblock \bibinfo{title}{Can Large Language Models Faithfully Express Their Intrinsic Uncertainty in Words?}
\newblock
\showeprint[arxiv]{2405.16908}~[cs.CL]
\urldef\tempurl%
\url{https://arxiv.org/abs/2405.16908}
\showURL{%
\tempurl}


\bibitem[Zaharia et~al\mbox{.}(2024)]%
        {compound-ai-blog}
\bibfield{author}{\bibinfo{person}{Matei Zaharia}, \bibinfo{person}{Omar Khattab}, \bibinfo{person}{Lingjiao Chen}, \bibinfo{person}{Jared~Quincy Davis}, \bibinfo{person}{Heather Miller}, \bibinfo{person}{Chris Potts}, \bibinfo{person}{James Zou}, \bibinfo{person}{Michael Carbin}, \bibinfo{person}{Jonathan Frankle}, \bibinfo{person}{Naveen Rao}, {and} \bibinfo{person}{Ali Ghodsi}.} \bibinfo{year}{2024}\natexlab{}.
\newblock \bibinfo{title}{The Shift from Models to Compound AI Systems}.
\newblock \bibinfo{howpublished}{\url{https://bair.berkeley.edu/blog/2024/02/18/compound-ai-systems/}}.
\newblock


\bibitem[Zhang et~al\mbox{.}({[n.\,d.]})]%
        {zhang2023building}
\bibfield{author}{\bibinfo{person}{Hongxin Zhang}, \bibinfo{person}{Weihua Du}, \bibinfo{person}{Jiaming Shan}, \bibinfo{person}{Qinhong Zhou}, \bibinfo{person}{Yilun Du}, \bibinfo{person}{Joshua Tenenbaum}, \bibinfo{person}{Tianmin Shu}, {and} \bibinfo{person}{Chuang Gan}.} \bibinfo{year}{[n.\,d.]}\natexlab{}.
\newblock \showarticletitle{Building Cooperative Embodied Agents Modularly with Large Language Models}. In \bibinfo{booktitle}{\emph{NeurIPS 2023 Foundation Models for Decision Making Workshop}}.
\newblock


\end{thebibliography}

%TC:ignore

\appendix
\break
\section{An Example Plan from Planning Agent}\label{app:dag-plan}
\lstset{basicstyle=\ttfamily}
\begin{tcolorbox}[
% title=DAG-formatted plan example, 
label={app:planning_prompt}, 
boxsep=2pt,left=3pt,right=5pt,top=0pt,bottom=0pt,
width=\linewidth]
\scriptsize
\begin{lstlisting}
{
    "id_": 0,
    "question": "Janet's ducks lay 16 eggs per day. 
    She eats three for breakfast every morning and 
    bakes muffins for her friends every day with four. 
    She sells the remainder at the farmers' market daily 
    for $2 per fresh duck egg. How much in dollars does 
    she make every day at the farmers' market?",
    "answer": "Janet sells 16 - 3 - 4 = <<16-3-4=9>>9 
    duck eggs a day.\nShe makes 9 * 2 = $<<9*2=18>>18 
    every day at the farmers' market.\n#### 18",
    "system_prompt": "You are a helpful assistant in 
    solving math questions",
    "user_prompt": {
        "1": "Identify the number of eggs laid per day, 
        eggs eaten for breakfast, eggs used for 
        baking, and the price per egg from the question.",
        "2": "Subtract the number of eggs eaten for 
        breakfast from the number of eggs laid per day.",
        "3": "Subtract the number of eggs used for 
        baking from the remaining eggs after breakfast.",
        "4": "Multiply the number of eggs available for 
        sale by the price per egg."
    },
    "plan": [
        {
            "id": 1,
            "name": "identify operands",
            "input": "question",
            "output": "number of eggs laid per day, 
            eggs eaten for breakfast, eggs used for 
            baking, price per egg"
        },
        {
            "id": 2,
            "name": "subtract",
            "input": "number of eggs laid per day, 
            eggs eaten for breakfast",
            "output": "remaining eggs after breakfast"
        },
        {
            "id": 3,
            "name": "subtract",
            "input": "remaining eggs after breakfast, 
                eggs used for baking",
            "output": "eggs available for sale"
        },
        {
            "id": 4,
            "name": "multiply",
            "input": "eggs available for sale, 
            price per egg",
            "output": "total earnings from selling eggs"
        }
    ],
    "edges": [
        [1, 2],
        [2, 3],
        [1, 3],
        [3, 4],
        [1, 4]
    ]
}
\end{lstlisting}
\end{tcolorbox}

\section{Agent Verifier Features}
\label{app:verifier-features}
We collect features such as uncertainty metrics from LLM agents' internals during execution, consistency-related metrics from additional LLM prompting, criteria-based scores from external LLM judges, and node characteristics features. We use GPT-4o for all experiments, employing a temperature setting of 0.7 for consistency-related metrics to obtain as diverse answers as possible, and a temperature of 0.1 for the remaining experiments ~\citep{xiong2023can}.
%%%%%%%%%%%%%%%%%
% We list our features below.
% While shared features mainly represent the model confidence in performing the execution, the subtask-specific features are derived from subjective criteria pre-designed per subtask.

\subsection{Features on Agent Uncertainty}\label{app:verifier-features-uncertainty}
%We extract numeric features from the execution results that represent uncertainty of each executor module executed, in which the LLMs are considered to output definitive responses\todo{cite temperature assigning paper}. The shared features are described as follows:
%\todo{add citation for each feature}
\begin{itemize}
    \item{\textbf{Verbalized confidence}~\cite{xiong2023can} reflects how confident the executor module is about its output, often derived from explicit confidence scores or qualitative indications of certainty (e.g., ``0.7'').}
    \item{\textbf{Logit-based confidence}~\cite{huang2023lookleapexploratorystudy} reflects the average of the exponentials of the token log probabilities (LP): 
        \begin{align}
        LP_{avg}
        = \frac{1}{N} \sum_{i=1}^{N} p(s_j \mid x_i)
        = \frac{1}{N} \sum_{i=1}^{N} \exp(\log p(s_j \mid x_i)), 
        \end{align}
        where \( N \) is the number of tokens and $\log p(s_j \mid x) = \sum_{i=1}^{j} \log p(s_i \mid s_{<i})$, where \( s_i \) is the \( i \)-th output token and \( s_{<i} \) denotes the set of previous tokens. We denote $\log p(s_j \mid x_i)$ as token log probability.
}
    \item{\textbf{Softmax-based confidence} reflects the average of softmax values across the generated tokens, providing a measure of the overall uncertainty of the model based on the top-k token log probabilities: 
        \begin{equation}
        Softmax_{avg} = \frac{1}{N} \sum_{n=1}^{N} \frac{\exp(\mathbf{z}_{n})}{\sum_{i=1}^{k} \exp(\mathbf{z}_{n,i})},
        \end{equation}
    where \( \mathbf{z}_n \) is the token log probability and \( N \) is the number of tokens.
    }
    \item{\textbf{Entropy-based confidence}~\cite{huang2023lookleapexploratorystudy} reflects the average of entropy values across the generated tokens, providing a measure of the overall uncertainty of the model based on the top-k token log probabilities:
    \begin{equation}
    Entropy_{avg} = \frac{1}{N} \sum_{n=1}^{N} \left( -\sum_{i=1}^{k} \frac{\exp(\mathbf{z}_{n,i})}{\sum_{j=1}^{k} \exp(\mathbf{z}_{n,j})} \log\left(\frac{\exp(\mathbf{z}_{n,i})}{\sum_{j=1}^{k} \exp(\mathbf{z}_{n,j})}\right) \right),
    \end{equation}
    where \( \mathbf{z}_n \) is the token log probability and \( \mathbf{z}_{n,i} \) is the \( i \)-th top log probability.
    }
    \item{Features from the a separate LLM evaluator capture the uncertainty in its assessment of the initial LLM execution.}
            \begin{itemize}[label=--]
                \item{\textbf{Verbalized confidence from external LLM evaluator} is attained directly from the external evaluator's generated response.}
                \item{\textbf{Logit-based confidence from LLM evaluator} is attained by the exponential of the logit value of the LLM evaluator\'s verification assessment (e.g., logit value of the \texttt{TRUE}).}
            \end{itemize}
    \item{Features using self-consistency~\cite{wang2022self} technique; we run the same prompt five times and aggregate the confidence values in the following ways:}
        \begin{itemize}[label=--]
        \item{\textbf{Self-consistency (Type A): frequency }~\cite{yona2024largelanguagemodelsfaithfully} measures the confidence of the executor module by the degree of agreement among the candidate outputs and integrates the inherent uncertainty in the model's output~\citep{xiong2023can}:
            \begin{equation}
            \text{Confidence}_{freq} = \frac{1}{M} \sum_{i=1}^{M} \mathbb{1}_{\left\{ \text{agreement}(\hat{Y}_i, \tilde{Y}) > \theta \right\}},
            \end{equation}
            where \( \mathbb{1}_{\left\{ \text{condition} \right\}} \) is the indicator function that returns 1 if the candidate answer \( \hat{Y}_i \) is consistent with the initial execution result \( \tilde{Y} \) based on the agreement threshold \( \theta \), and 0 otherwise. Here, \( M \) denotes the number of candidate answers, and \( \theta \) is the threshold for agreement. We use the answer equivalence package PEDANT~\citep{li2024panda} with their recommended threshold of 0.5 to assess the agreement.
        }
        \item{\textbf{Self-consistency (Type B): verbalized confidence }~\cite{xiong2023can} measures the average verbalized confidence among the subset of candidate answers identified as correct.
            \begin{equation}
            \text{Confidence}_{verb}= \frac{\sum_{i=1}^{M} C_i^{verb} \cdot \mathbb{1}_{\{\text{correctness}_i = 1\}}}{\sum_{i=1}^{M} \mathbb{1}_{\{\text{correctness}_i = 1\}}},
            \label{equation:self_con_verb}
            \end{equation}
            where \( M \) is the total number of candidate answers and \( C_i \) denotes the verbalized confidence of candidate answers.
        }
        \item{\textbf{Self-consistency (Type C): logit-based } measures the average logit-based confidence among the subset of candidate answers identified as correct. 
            \begin{equation}
            \text{Confidence}_{log}= \frac{\sum_{i=1}^{M} C_i^{log} \cdot \mathbb{1}_{\{\text{correctness}_i = 1\}}}{\sum_{i=1}^{M} \mathbb{1}_{\{\text{correctness}_i = 1\}}},
            \label{equation:self_con_logit}
            \end{equation}
            where \( M \) is the total number of candidate answers and \( C_i \) denotes their average log probabilities.
        }
    \end{itemize}

\end{itemize}

\subsection{Features on Agent-Specific Criteria}\label{app:verifier-features-criteria}
\label{feature_critera}

\begin{itemize}
    \item{\textbf{Per-criteria scores by LLM judges} These features are introduced to measure the successfulness from human perspective (details in Table \ref{tab:expanded_agent_criteria}).
    We prompt an LLM to score the execution results based on these predefined criteria per subtask. For example, the execution result of an ``add'' excutor, "9 apples", might be evaluated with the following binary scores: \{"accuracy of numerical value": 1.0, "sufficiency of context information": 0.0, "adherence to format": 1.0\}. By assessing the correctness of the LLM executor's outputs and using these evaluations as features in training a verification agent, we ensure that the agent's detection is grounded in reliable, human-aligned guidelines.\footnote{Given that each subtask has a varying number of criteria, we create a one-hot vector to indicate the specific subtask to which each sample refers. This vector is then concatenated with a matrix containing the union of criteria columns across all subtasks, populated with the corresponding binary values from the execution result.}}
\end{itemize}

\subsection{Other Features}\label{sec:plan_features}

\begin{itemize}
    \item{\textbf{Subtask type} is represented as one-hot encoding for all subtasks in our taxonomy.}
    \item{\textbf{Features on Plan Structures}}
    \begin{itemize}
        \item{\textbf{Number of preceding subtasks} is the number of previous subtasks that this subtask is depending on (i.e., in-degree). We incorporate this feature to reflect the dependency information between subtasks within the plan.}
        \item{\textbf{Source distance} measures the shortest chain length to reach the current subtask as a proxy for node importance.}
    \end{itemize}
\end{itemize}

\section{List of Prompts}
\label{app:prompts}

We provide the prompts used for planning, agent execution, and criteria evaluation by external evaluators.

\begin{promptbox}{Prompt used for planning} \label{app:planning_prompt}
\scriptsize
You are a planner responsible for creating high-level plans to solve any tasks using a set of agents.\\
Your goal is to break down a given task into a sequence of subtasks that, when executed correctly
by the appropriate agents, will lead to the correct solution.\\
 
For each step in the plan: \\
1. Describe the subtask the agent must perform.\\
2. Provide a brief, self-contained description of the expected inputs and outputs. Do not include any specific values or examples.\\
3. Provide a user prompt for each task that includes the expected input and output information.\\

Represent your plan as a graph where each node corresponds to a step, and each edge represents a dependency between two steps.\\
If a node requires the output from a previous node as an input, ensure it is included in the edge list.\\
The output should be structured in the following JSON format:\\
$\{$ \\
    ``nodes'': <list of JSON nodes $\{$ ``id'': <node id as integer>, ``name'': <assigned agent name>, ``task'': <task instruction>, \\ \hspace{5cm}``input'': <list of inputs>, ``output'': <list of outputs>$\}$>,\\
    ``edges'': <list of tuples [node\_id, node\_id]>\\
    ``user\_prompts'':  <list of strings per node>\\
$\}$\\
Available agents:
\{\texttt{agent taxonomy}\}\\

Examples\\
\{\texttt{plan demonstration examples}\}

\tcblower
\scriptsize
\{\texttt{task query}\}
\end{promptbox}

\begin{promptbox}{Prompt used for agent execution and verbalized confidence}\label{app:execution_prompt}
\scriptsize
Use the following contextual information to answer: \{\texttt{context info}\}. \\
If contextual information is ``None'', answer it without external information. \\
 JUST PERFORM WHAT YOU ARE ASKED TO DO, DO NOT ANSWER THE QUESTION, JUST BECAUSE THE QUESTION EXISTS IN THE PROMPT. \\
Your answer should always be in JSON object format. 
\{answer: <answer>, confidence: <confidence>\}.

\tcblower
\scriptsize
\{\texttt{subtask instruction prompt from the plan}\} + Also, provide how confident you are in your answer.\\
If not, use your own memory to execute the prompt as best as you can. \\
If you do not know the answer, your confidence should be 0.0. \\
The answer format should be like \{answer: <text>, confidence: <float value between [0-1]>\}.
\end{promptbox}

\begin{promptbox}{Prompt used for using LLM evaluator with human-defined criteria}
\scriptsize
You're a helpful assistant that evaluates an agent \{agent\}'s answer in different criteria.\\          
Your answer should always be in JSON format.\\
\{`criteria':`criteria score'\}." 

\tcblower
\scriptsize
Please evaluate the following agent's answer to a user prompt with the following context information.\\
        If the context information is `None', ignore and use your own knowledge to answer.\\
        Here are some examples to help you score the agent's answer: \{agent examples\}\\
        The user prompt: \{user\}.\\
        The context information: \{context info\}.\\
        The agent: \{agent\}. \\
        The agent's input format: \{agent input\}.\\
        The agent's output format: \{agent output\}.\\
        The agent's answer: \{answer\}.\\
        You should find the agent's essential criteria to evaluate the answer from \{agent criteria list\}\\
        Then, score each criterion in a float value between [0-1] in the 'criteria score' placeholder.\\
        Your answer should look like:\\
        \{'criteria score': \{'<criteria 1>': <float value between [0-1]>, '<criteria 2>': <float value between [0-1]>, '<criteria 3>': <float value between [0-1]>, '<criteria 4>': <float value between [0-1]>, '<criteria 5>': <float value between [0-1]>\}
        \}\\
        Your answer should always be in JSON object format: \{'criteria':'criteria score'\}.
\end{promptbox}

% \section{Agent Registry}
\section{Agent Registry Curation using Chain-of-Thought Prompting}\label{app:cot}
We provide the AI practioners to examine 50 samples resulting from one-shot promoting, and determine which agents are adequate for math reasoning application. As an example, we provide result for the task in Figure~\ref{fig:agent_failure_example}:

\begin{exbox}{CoT planning result example}
% \begin{tcolorbox}[title=CoT planning result example, label={app:planning_prompt}, width=\linewidth,
% boxsep=2pt,left=3pt,right=5pt,top=0pt,bottom=0pt]
\scriptsize
\begin{lstlisting}
{
    "question": "Janet's ducks lay 16 eggs per day.
    She eats three for breakfast every morning and 
    bakes muffins for her friends every day with four. 
    She sells the remainder at the farmers' market daily 
    for \$2 per fresh duck egg. How much in dollars 
    does she make every day at the farmers' market?",
    "answer": "Janet sells 16 - 3 - 4 = <<16-3-4=9>>9 
    duck eggs a day.\nShe makes 9 * 2 = $<<9*2=18>>18 
    every day at the farmer's market.\n#### 18",
    "result": {
        "role": "assistant",
        "plans": {
            "step1": "Determine the total number of eggs 
            Janet uses for breakfast and baking muffins.",
            "step2": "Subtract the number of eggs used 
            for breakfast and baking muffins from the 
            total number of eggs laid per day to find 
            the number of eggs she sells.",
            "step3": "Calculate the total revenue by
            multiplying the number of eggs she sells 
            by the price per egg."
        },
        "answers": {
            "step1": "Janet uses 3 eggs for breakfast 
            and 4 eggs for baking muffins, so she uses 
            a total of 3 + 4 = 7 eggs per day.",
            "step2": "Janet's ducks lay 16 eggs per day. 
            After using 7 eggs, she has 16 - 7 = 9 eggs 
            left to sell.",
            "step3": "Janet sells 9 eggs at $2 per egg, 
            so she makes 9 * $2 = $18 per day at 
            the farmers' market."
        }
    }
}
\end{lstlisting}
% \end{tcolorbox}
\end{exbox}

\section{Human-Defined Criteria for Agents}
\label{app:human-criteria}

Table~\ref{tab:expanded_agent_criteria} lists evaluation criteria for each agent in our agent registry.

\section{Crowdsourced Annotation Procedure}\label{app:user_study}

\begin{figure*}
  \includegraphics[width=\linewidth]{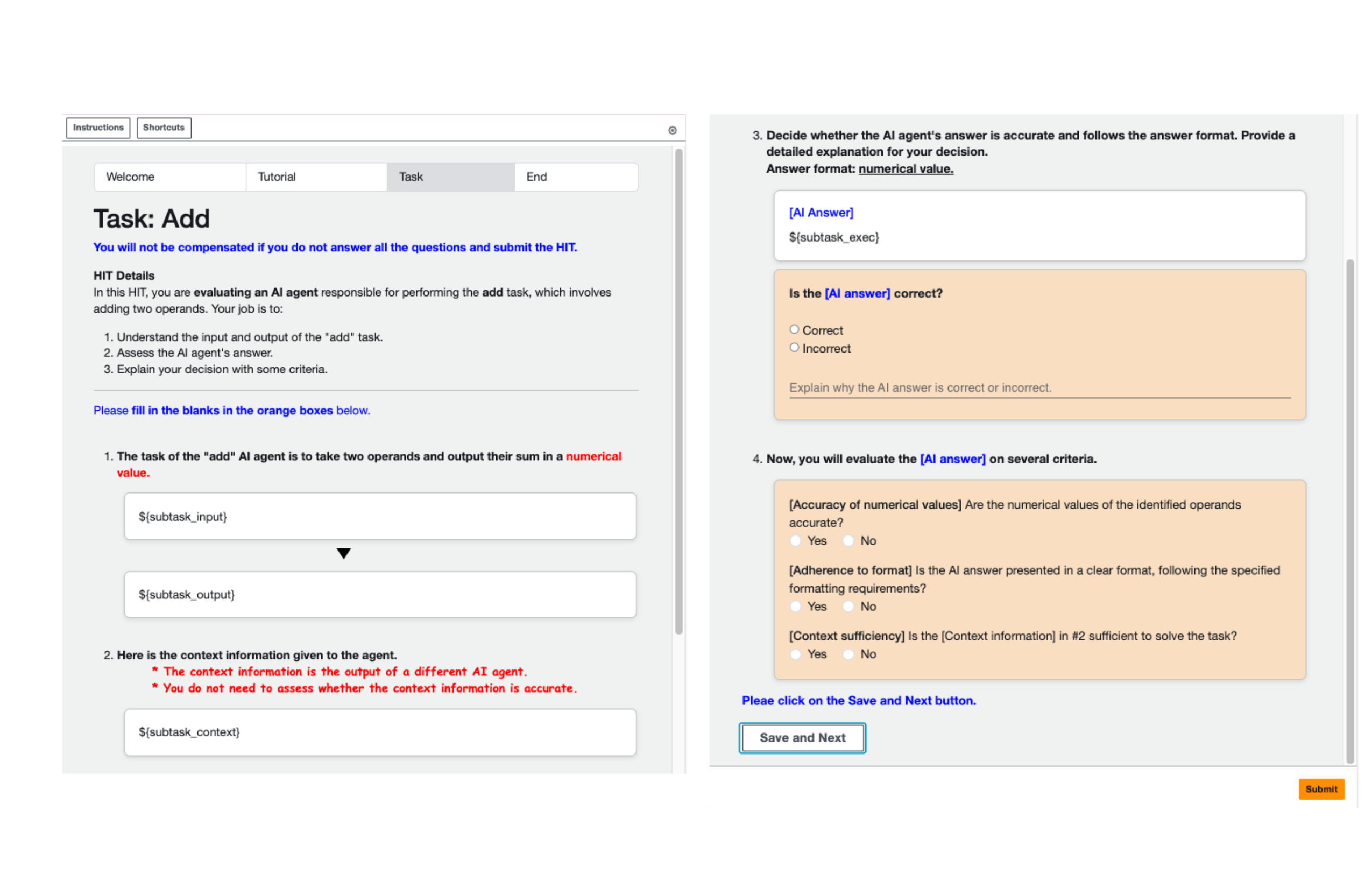}
  \caption{Example annotation for evaluation of LLM execution result of ``add'' subtask. We prohibit the users from moving on to the next page if they did not get the answer correct for questions in the tutorial.}
  \label{fig:actual_add}
\end{figure*}

We used MTurk platform to recruit crowdworkers for labeling the GSM8K dataset. For each instance, three labels were collected from workers who passed the qualification test, with an average pay rate of \$14 per hour. 
For a subtask assigned to an agent, workers were asked to annotate whether the agent's answer was successful or not, given its role, input, and evaluation criteria. 
% Also, they indicate whether the answer satisfied the provided criteria.
A sample interface illustrating the annotation process is shown in Figure~\ref{fig:actual_add}.

\section{Experiment Details}

\subsection{Dataset Statistics}\label{subtask_stat}
We partition each dataset to evaluate our approach. First, we reserve a held-out test set of tasks (math questions) used exclusively to evaluate the performances of aggregation metrics (Table~\ref{tab:task_task_train_split}). The remaining instances, which are decomposed into subtasks, are further split into train and test sets for training agent verifiers (Table~\ref{tab:subtask_train_test_split}). This hierarchical split ensures that agent verifiers are validated independently of the aggregator.

\begin{table}[htp]
\caption{Numbers of tasks in train-test split for aggregation.}
\centering
\vspace{-1em}
\begin{tabular}{lcc}
\toprule
\textbf{Dataset} & \textbf{Train} & \textbf{Test} \\
\midrule
GSM8K               & 444 & - \\
Date Understanding  & 137 & 35 \\
Object Counting     & 158 & 40 \\
Arithmetic Multistep & 123 & 31 \\
% \midrule
% \textbf{Total}      & 862 & 106 \\
\bottomrule
\end{tabular}
\vspace{-1em}
\label{tab:task_task_train_split}
\end{table}

\begin{table}[htp]
\caption{Numbers of subtasks in train-test split for agent verifiers.}
\centering
\vspace{-1em}
\begin{tabular}{lcc}
\toprule
\textbf{Dataset} & \textbf{Train} & \textbf{Test} \\
\midrule
GSM8K               & 778 & 195 \\
Date Understanding  & 327 & 82 \\
Object Counting     & 211 & 53 \\
Arithmetic Multistep & 578 & 145 \\
% \midrule
% \textbf{Total}      & 1894 & 475 \\
\bottomrule
\end{tabular}
\vspace{-1em}
\label{tab:subtask_train_test_split}
\end{table}

\subsection{Agent Verifier Accuracy Using Different Models}\label{app:different_models}

\begin{figure}[th]
  \includegraphics[width=\linewidth]{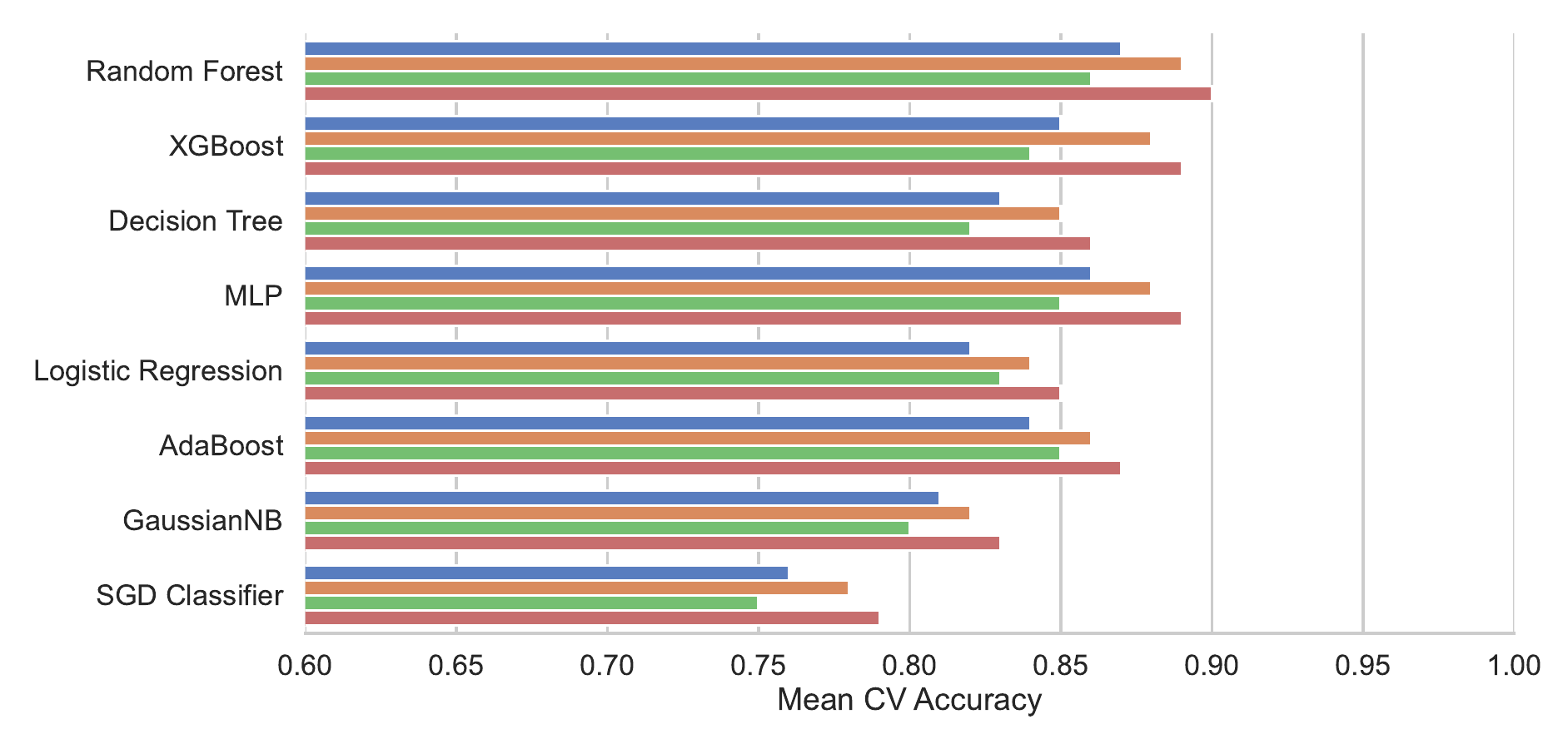}
  \caption{Verifier accuracy per model across datasets: average over 5-fold cross-validation.}
  \Description{\todo{add}}
  \label{fig:sub_verifier_each}
\end{figure}

We train agent verifiers separately for each dataset using the following machine learning models: Logistic Regression~\citep{cox1958regression}, SGD Classifier~\citep{bottou2010large}, Decision Tree~\citep{breiman1986classification}, Random Forest~\citep{parmar2019review}, AdaBoost~\cite{Schapire2013}, XGBoost~\citep{freund1997decision}, Gaussian Naive Bayes (GaussianNB)~\citep{john1995estimating}, and Multi-Layer Perceptron (MLP)~\citep{rumelhart1986learning}.
%\todo{cite individually}
% Fig.~\ref{fig:sub_verifier_each} represents the accuracy of the verifier models, with results averaged over 5-fold cross-validation.
Among all tested models, the random forest classifier with 100 tree estimators achieved the highest average accuracy among other models in four datasets (0.88).
Figure~\ref{fig:sub_verifier_each} represents the accuracy of the verifier models, with results averaged over 5-fold cross-validation.
Among the tested models, the random forest classifier with 100 tree estimators achieved the highest overall accuracy in four datasets.

\renewcommand{\arraystretch}{1.5} % Increase row spacing for better readability

\begin{table*}[p]
\centering
\label{tab:expanded_agent_criteria}
\small
\begin{adjustbox}{width=\textwidth} 
\begin{tabular}{@{}p{9cm}p{10cm}@{}}
\toprule
\textbf{Subtask and Description} & \textbf{Essential Criteria} \\
\midrule

Identify Operands — Identify operands with text description of each operand & 
\textbf{Accuracy}: Are numerical values accurate? \newline
\textbf{Relevance}: Are all operands relevant? \newline
\textbf{Coverage}: Are all necessary operands identified? \newline
\textbf{Clarity}: Are operand descriptions clear? \newline
\textbf{Format Adherence}: Is the output correctly formatted? \\

\midrule

Add — Add numbers or dates & 
\textbf{Accuracy}: Is the sum correct? \newline
\textbf{Format Adherence}: Is the output correctly formatted? \newline
\textbf{Context Sufficiency}: Is the context enough to solve the task? \\

\midrule

Subtract — Subtract numbers or dates & 
\textbf{Accuracy}: Is the result correct? \newline
\textbf{Format Adherence}: Is the output correctly formatted? \newline
\textbf{Context Sufficiency}: Is the context enough to solve the task? \\

\midrule

Multiply — Multiply numbers & 
\textbf{Accuracy}: Is the result correct? \newline
\textbf{Format Adherence}: Is the output correctly formatted? \newline
\textbf{Context Sufficiency}: Is the context enough to solve the task? \\

\midrule

Divide — Divide numbers & 
\textbf{Accuracy}: Is the result correct? \newline
\textbf{Format Adherence}: Is the output correctly formatted? \newline
\textbf{Context Sufficiency}: Is the context enough to solve the task? \\

\midrule

Filter — Filter a list based on a condition & 
\textbf{Relevance}: Are irrelevant items excluded? \newline
\textbf{Completeness}: Are all valid items included? \newline
\textbf{Format Adherence}: Is the output correctly formatted? \newline
\textbf{Context Sufficiency}: Is the context enough to solve the task? \\

\midrule

Sort — Sort a list by an attribute & 
\textbf{Correctness}: Is the order accurate? \newline
\textbf{Completeness}: Are all items included? \newline
\textbf{Format Adherence}: Is the output correctly formatted? \newline
\textbf{Context Sufficiency}: Is the context enough to solve the task? \\

\midrule

Convert Format — Convert input from one format to another & 
\textbf{Accuracy}: Was the conversion correct? \newline
\textbf{Format Adherence}: Is the output correctly formatted? \newline
\textbf{Context Sufficiency}: Is the context enough to perform the conversion? \\

\midrule

Date Lookup — Identify year, month, and day from a natural language description & 
\textbf{Accuracy}: Is the date correctly identified? \newline
\textbf{Format Adherence}: Is the output correctly formatted? \newline
\textbf{Context Sufficiency}: Is the context enough to extract the date? \\

\bottomrule

\end{tabular}
\end{adjustbox}
\caption{Human-designed agent criteria. Each agent's criteria are assigned by users based on their own experience performing the task using the agent registry. Thus, these criteria are grounded in human needs and are integrated into LLM evaluators, with their outputs used as part of our agent verifier features.}
\end{table*}

%TC:endignore
%\label{sec:appendix}
%\input{tex/90_appendix}
\end{document}